\ifdefined\pdfminorversion\pdfminorversion=7\relax\fi
\documentclass[manuscript,screen]{acmart}
\acmJournal{TIST}
\setcopyright{none}
\acmDOI{}
\acmVolume{}
\acmNumber{}

\usepackage{amsmath}
\usepackage{booktabs}
\usepackage{graphicx}
\usepackage{placeins}

\begin{document}
\title[TopologyUnderstandingNet]{TopologyUnderstandingNet (TUN): Geometry-Conditioned Classification of Persistence Diagram Points}

\author{Yu Chen}
\email{chenyu.math@zju.edu.cn}
\affiliation{\department{School of Mathematical Sciences}
  \institution{Zhejiang University}\city{Hangzhou}\country{China}}
\author{Hongwei Lin}
\email{hwlin@zju.edu.cn}
\affiliation{\department{School of Mathematical Sciences}
  \institution{Zhejiang University}\city{Hangzhou}\country{China}}

\begin{abstract}
Persistence diagrams (PDs) provide a compact description of the topology underlying a point cloud. However, identifying which diagram entries correspond to features of the source shape remains challenging, limiting their use in downstream geometric tasks. In this paper, we study automatic significance detection for degree-one persistence diagrams and propose TopologyUnderstandingNet (TUN), a multimodal network for classifying individual diagram entries. TUN combines enhanced birth–death descriptors and self-attention with a PointNet-style point-cloud encoder. A geometry-conditioned correction module adjusts the decision threshold and individual entry scores, allowing the predicted significant set to depart from lifetime ranking. We evaluate TUN on four geometric categories using reviewed entry-level labels. It achieves higher F1 averaged across categories than classical selection rules and PD-only set models. We further evaluate the selected intervals through CAD source-identity retrieval. At matched selection counts, TUN improves retrieval over lifetime ranking for queries with added outliers, demonstrating the downstream value of selecting diagram entries beyond persistence order.
\end{abstract}

\maketitle

\section{Introduction}
Persistent homology tracks connected components, loops, and voids across scale and records their birth and death values in a persistence diagram (PD)~\citep{edelsbrunner2010computational}. Point clouds capture geometry in applications such as robotic mapping~\citep{zhang2014loam}, and persistent homology is used to study structure in medical images~\citep{singh2023topological} and proteins~\citep{xia2014persistent}. Interpreting a diagram computed from a sample requires deciding which entries capture features of the source.

A threshold on persistence selects intervals in order of lifetime~\citep{fasy2014confidence}. Even a threshold learned separately for every diagram~\citep{kudeshia2026topogat} preserves this order within a diagram. The distinction matters when a small loop is present in the source shape but a longer-lived loop forms only as the shape thickens. Selecting more or fewer entries by lifetime alone cannot retain the shorter interval while rejecting the longer one.

We formulate the choice of intervals as supervised classification in a fixed $H_1$ alpha-filtration diagram (1-PD)~\citep{edelsbrunner2010computational}. We initialize the positive set with the longest intervals up to the source loop count, then review their identities against the source geometry. Throughout the paper, a positive or \emph{significant} entry denotes membership in this reviewed set. Different geometries can give rise to the same diagram~\citep{bubenik2020persistent}, which motivates comparing predictors with and without access to the sampled point cloud.

\begin{figure}[htbp]
    \centering
    \includegraphics[width=0.9\linewidth]{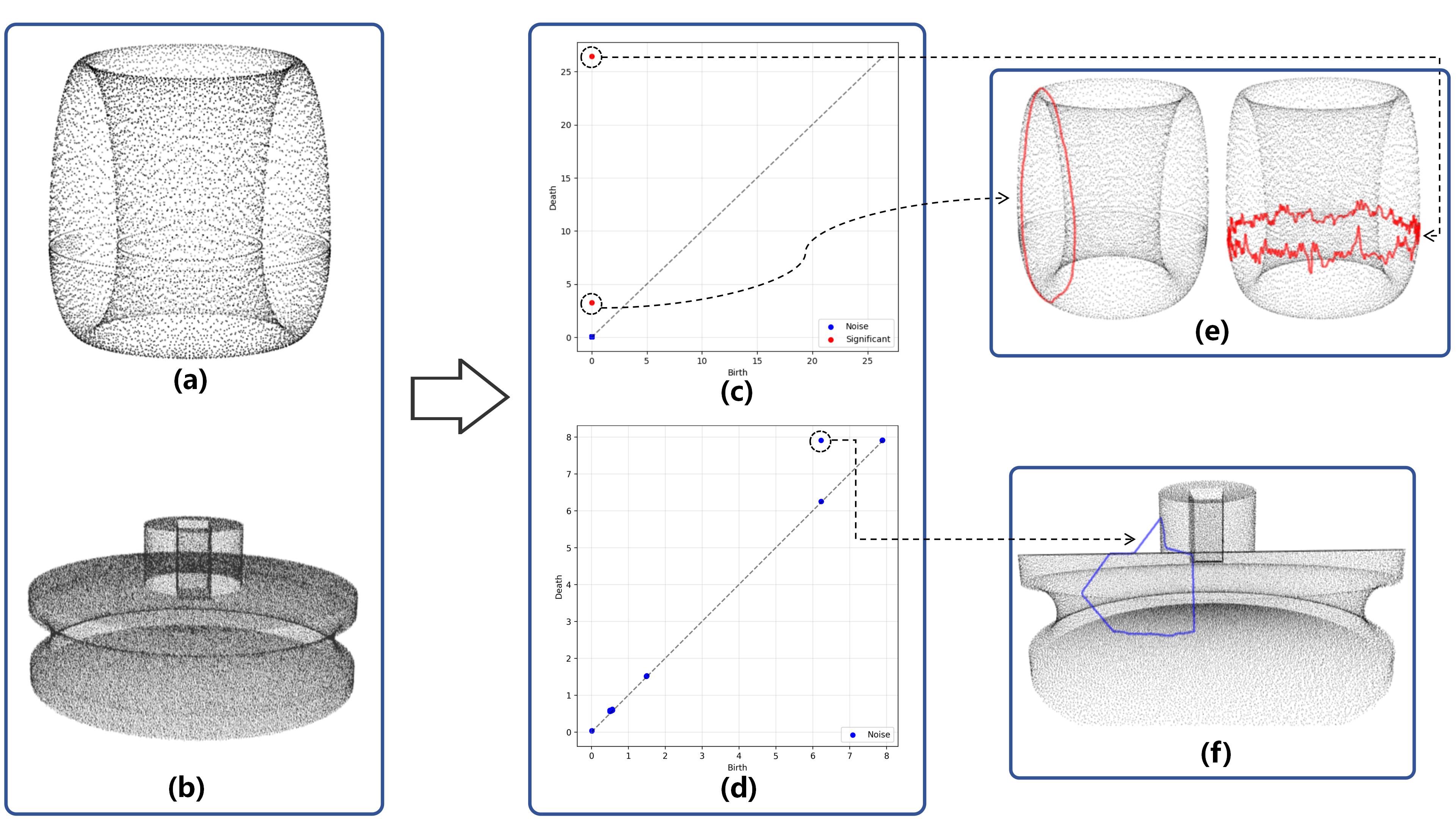}
    \caption{An illustration of the operational point-labeling task in 1-PDs. (a) and (b): point clouds sampled from CAD models with first Betti numbers 2 and 0. (c) and (d): the corresponding 1-PDs and benchmark labels, where red and blue denote positive and negative entries. (e): volume-optimal representative cycles for the two positive entries in (c), both supported near loops of the sampled shape. (f): a representative cycle for the most persistent negative entry in (d); much of the cycle lies away from the sampled surface.}
    \Description{Two CAD point clouds appear next to their birth--death plots, where red entries are positive and blue entries negative. The first cloud carries two red representative cycles. A blue cycle for the second cloud extends away from its surface.}
    \label{fig:illustrate}
\end{figure}

Figure~\ref{fig:illustrate} shows volume-optimal representative cycles~\citep{obayashi2018volume} for three reviewed entries. The two positive entries in panel (c) give cycles near loops of the sampled shape. The cycle for the long-lived negative entry in panel (d) extends away from the surface. Appendix~\ref{app:benchmark} includes a further example of visually similar diagrams with different reviewed labels.

We introduce \textbf{TopologyUnderstandingNet (TUN)} for this task. The model first scores each entry in the context of its diagram. A shared point-cloud descriptor then shifts the decision threshold and supplies a gated correction specific to each entry.

The work makes three contributions.
(1) We construct a four-category benchmark with reviewed labels for individual $H_1$ entries and give sufficient conditions for early-birth selection to identify source loops.
(2) We develop a supervised predictor whose entry scores can depart from lifetime order within a diagram.
(3) Across ten runs, TUN reaches an unweighted four-category mean F1 of $0.9965$, compared with $0.9933$ for DeepSets. Across nine CAD query corruptions, mean average precision is $0.3540$ for TUN and $0.3215$ for lifetime ranking at the same per-observation counts. The largest gains occur under added outliers, where the selected intervals help identify an independently sampled source mesh.

\section{Related Work}\label{sec:related_work}

\subsection{Learning from topological descriptors}

Topological data analysis studies the shape of data through constructions such as persistent homology~\citep{edelsbrunner2008persistent,edelsbrunner2010computational,zomorodian2004computing,wasserman2018topological}. Prior work covers algorithms for several filtration families~\citep{edelsbrunner2010computational,maria2014gudhi}, stability under perturbations~\citep{cohen2005stability}, and applications in molecular science~\citep{xia2014persistent,bramer2020atom}, materials science~\citep{hiraoka2016hierarchical,obayashi2022persistent}, geometric design and processing~\citep{bruel2020topology,dong2022topology,he2023robust,chen2025robust}, and medical image processing~\citep{brito2025persistent,singh2023topological}.

For terrain imagery, \citet{he2022terrain} use a contour tree to model topological dependencies between pixel labels in flood mapping. TDA has also been combined with machine learning for feature extraction and classification~\citep{pun2022persistent,barnes2021comparative}. PersLay~\citep{carriere2020perslay} maps PDs to vector representations through learnable functions. Persformer~\citep{reinauer2021persformer} instead treats each diagram as an unordered point set and applies a Transformer without fixed vectorization. For point clouds, TopologyNet learns a direct map from geometry to persistence images~\citep{zhou2022topologynet}, whereas TopoLayer encapsulates PD vectorization in a module that augments point-cloud backbones~\citep{guan2025topolayer}. TOPF constructs node-level topological features on the source point cloud from algebraic-topological and differential-geometric quantities~\citep{grande2025topf}. These methods produce diagram-level or source-point features. For our entry-level classification task, set-processing principles motivate the DeepSets and Set Transformer baselines used in our evaluation~\citep{zaheer2017deep,lee2019set}.
Other studies incorporate PDs into loss functions to impose topological constraints during training~\citep{wong2021persistent,hu2025topogen,jignasu2024sdfconnect}.

\subsection{Selecting significant diagram entries}

The significance of points in PDs has been studied using statistical and geometric approaches. 
A common heuristic ranks points by persistence and declares the top $k$ (or the single most persistent point) as significant, discarding the remainder as noise~\citep{bruel2020topology,dong2022topology}. 
A statistical alternative constructs a confidence set around the diagonal through bootstrap resampling~\citep{fasy2014confidence}. The result depends on the point-cloud density and user-selected parameters. Clustering-based strategies, such as $k$-means with $k=2$, partition PD points into significant and noise classes~\citep{he2023robust,chen2025robust}. A two-cluster partition cannot represent a diagram in which all retained entries belong to the same class.
Other statistical approaches assess whether individual features are unusual or recur across samples. \citet{bobrowski2023null} use a null model for transformed death-to-birth ratios to test individual features. Cycle registration uses spatial location and persistence intervals to match cycles across samples~\citep{reani2023cycle}. Subsampling can also combine persistence summaries across samples~\citep{chazal2015subsampling}. Our target is the set of entries assigned to loops of one specified source shape.

TopoGAT~\citep{kudeshia2026topogat} augments ModelNet40 and ShapeNet with $H_1$ and $H_2$ Vietoris--Rips diagrams. Its training uses object-class labels and a loss on filtered diagrams. The predicted class probabilities and a diagram embedding determine one lifetime threshold per diagram and homology degree. Its increasing sigmoid mask preserves lifetime order within that diagram. TUN receives a target for each retained $H_1$ entry and can assign a higher score to a shorter, earlier-born interval than to a longer, later-born one.

Geometry can also change how the diagram is constructed. Distance-to-measure and kernel-distance filtrations~\citep{chazal2018robust} and Fermat distances~\citep{fernandez2023intrinsic} address robustness during filtration. Other approaches use latent graphs or landmarks~\citep{aupetit2005learning,gamble2010exploring}, or learn a filtration for a downstream task~\citep{nishikawa2023adaptive}. Our problem starts with an existing alpha diagram. Keeping its filtration fixed allows a predicted entry to refer to the same interval throughout the analysis.

\section{Preliminaries}\label{sec:preliminaries}

We define the alpha-filtration diagrams used in the entry-classification task, then describe two reference rules for selecting their entries.

\subsection{Alpha filtrations}

The convex hull of $n+1$ affinely independent points $\{u_0, u_1, \ldots, u_n\} \subset \mathbb{R}^N$ is an \textit{$n$-simplex}, denoted $[u_0, u_1, \ldots, u_n]$. Basic geometric primitives such as vertices, edges, triangles, and tetrahedra correspond to 0-, 1-, 2-, and 3-simplices, respectively. A \textit{simplicial complex} $K$ is a finite set of simplices satisfying two conditions: every face of a simplex in $K$ also belongs to $K$, and the intersection of any two simplices in $K$ is either empty or a shared face \citep{edelsbrunner2010computational}. The dimension of $K$ is the largest dimension of any simplex it contains. A simplicial complex $L$ is a \textit{subcomplex} of $K$ if $L \subseteq K$.

The \textit{alpha complex} is a simplicial complex constructed from a point cloud through Voronoi and Delaunay complexes. Given a point cloud $P=\{x_i\}_{i=1}^m \subset \mathbb{R}^n$, the \textit{Voronoi cell} for a point $x_i$ is the region of space closer to $x_i$ than to any other point in $P$:
\[V_{x_i} = \{ x\in \mathbb{R}^n: \|x-x_i\|\le \|x-x_j\|, \forall j\ne i \},\]
where $\|\cdot\|$ is the Euclidean norm. The nerve of the Voronoi cells defines the \textit{Delaunay complex}. It includes a simplex for every subset of points whose Voronoi cells have a non-empty common intersection:
\[\operatorname{Del}(P)=\left\{ [x_{i_{1}}, \cdots, x_{i_{q}}] : \bigcap_{j=1}^q V_{x_{i_j}} \neq \emptyset\right\}.\]

The alpha complex is a subset of the Delaunay complex controlled by a scale parameter $r \ge 0$. For each point $x \in P$, consider the intersection of its Voronoi cell $V_x$ with a closed ball of radius $r$ centered at $x$, denoted $R_x(r) = B_x(r) \cap V_x$. The \textit{alpha complex} $\operatorname{Alp}(P, r)$ consists of all simplices whose constituent points have intersecting restricted Voronoi regions:
\[
\operatorname{Alp}(P, r) = \left\{ [x_{i_{1}}, \dots, x_{i_{q}}] : \bigcap_{j=1}^q R_{x_{i_j}}(r) \neq \emptyset \right\}.
\]
The condition $R_x(r) \subseteq V_x$ ensures that $\operatorname{Alp}(P, r)$ is always a subcomplex of $\operatorname{Del}(P)$.

\begin{figure}[htbp]
    \centering
    \includegraphics[width=0.8\linewidth]{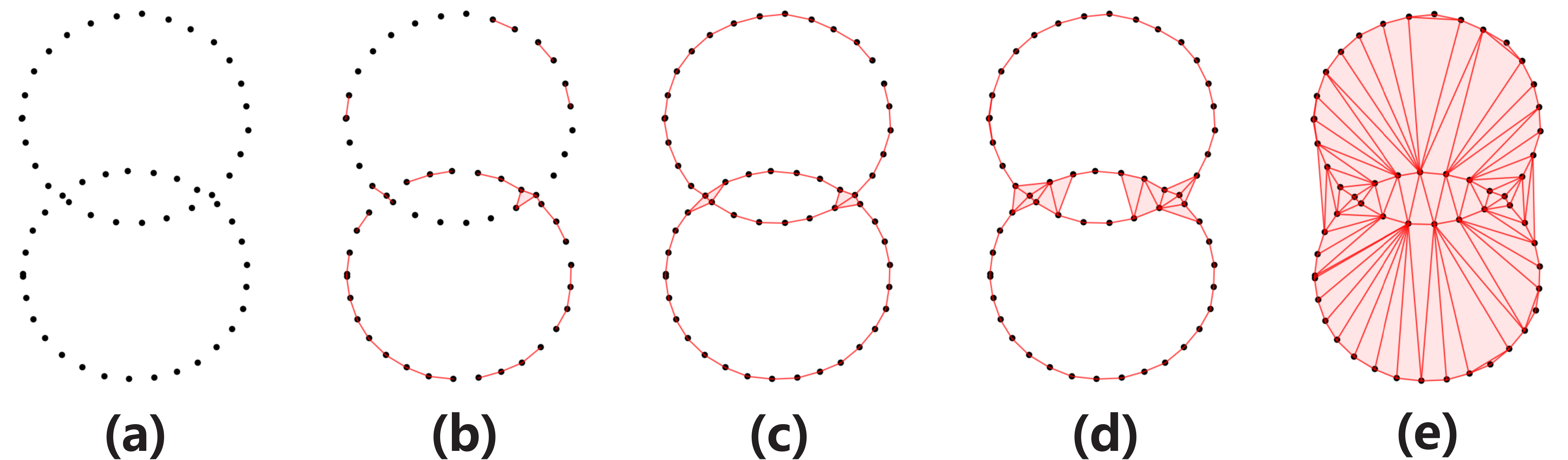}
    \caption{An alpha filtration. Additional simplices enter as the filtration parameter increases.}
    \Description{Five panels show the same points from left to right. Edges connect isolated vertices into loops, and triangular faces fill in parts of the complex as the radius grows.}
    \label{fig:alpha}
\end{figure}

Increasing the radius $r$ from zero generates a nested sequence of alpha complexes. This ordered family of complexes,
\[\operatorname{Alp}(P, r_0) \subseteq \operatorname{Alp}(P, r_1) \subseteq \cdots \subseteq \operatorname{Alp}(P, r_k)\subseteq \cdots,\]
where $0 \le r_0 < r_1 < \dots$, is known as the \textit{alpha filtration} of the point cloud $P$. Figure~\ref{fig:alpha} illustrates this process.

\subsection{Persistent homology and persistence diagrams}

\textit{Persistent homology} tracks topological features, such as connected components, tunnels, and voids, across a sequence of nested spaces~\citep{edelsbrunner2008persistent,edelsbrunner2010computational}. It records how long each feature persists over the filtration.

The formalism begins with the algebraic structure of a simplicial complex $K$. For a given dimension $n$, the set of all $n$-simplices in $K$ forms a basis for the \textit{$n$-chain group}, $C_n(K)$. An element of this group, an \textit{$n$-chain}, is a formal sum $c=\sum a_i\sigma_i$, where each $\sigma_i$ is an $n$-simplex and the coefficients $a_i$ are taken from a field, typically $\mathbb{Z}_2$ in computational settings. With coefficients in $\mathbb{Z}_2$, the \textit{boundary operator} $\partial_n: C_n(K) \to C_{n-1}(K)$ maps an $n$-simplex $\sigma=[x_0,x_1,\ldots,x_n]$ to the sum of its $(n-1)$-dimensional faces:
\[
\partial_{n}\sigma=\sum_{j=0}^{n} [x_{0},\ldots,\hat{x}_{j},\ldots,x_{n}],
\]
where $\hat{x}_j$ indicates that vertex $x_j$ is omitted. This operator defines two important subgroups of $C_n(K)$: the group of \textit{$n$-cycles}, $Z_n(K)=\operatorname{Ker}(\partial_n)$, consisting of chains with no boundary, and the group of \textit{$n$-boundaries}, $B_n(K)=\operatorname{Im}(\partial_{n+1})$, consisting of chains that bound $(n+1)$-chains. The \textit{$n$-th homology group} is the quotient group $H_{n}(K)=Z_{n}(K)/B_{n}(K)$, whose elements represent the $n$-dimensional holes of the complex.

For a filtration such as the alpha filtration described above,
\[K_0\subset K_1\subset \cdots \subset K_m,\]
the inclusion map $K_i \hookrightarrow K_j$ for $i \le j$ induces a homomorphism on the corresponding homology groups, $f_n^{i,j}: H_n(K_i)\to H_n(K_j)$. The $n$-th \textbf{persistent homology group} from $i$ to $j$ is the image of this homomorphism, $H_n^{i,j}=\operatorname{Im}f_n^{i,j}$. Its rank, $\beta_{n}^{i,j}=\operatorname{rank}H_{n}^{i,j}$, is the $n$-th \textbf{persistent Betti number}. Persistent homology therefore tracks which homology classes (features) ``persist'' as the filtration progresses and records the birth and death scales of their intervals.

Persistent homology is commonly summarized by a \textit{persistence diagram} (PD), a multiset of points in the birth--death plane with possibly infinite death times \citep{edelsbrunner2010computational}. For each dimension $n$, the $n$-th PD contains a point $(b_i, d_i)$ for each $n$-dimensional topological feature that appears at filtration value $b_i$ (its \textit{birth time}) and disappears at value $d_i$ (its \textit{death time}). The vertical distance from a point to the diagonal, $d_i - b_i$, defines its \textit{persistence}, which measures the feature's lifespan within the filtration.

Because a homology class is an equivalence class of cycles, a PD entry does not by itself specify a unique geometric curve. A \emph{representative cycle} selects one cycle from that class. Figure~\ref{fig:illustrate} uses volume-optimal representatives~\citep{obayashi2018volume} to illustrate the relation between the entries and the sampled geometry. Section~\ref{sec:datasets} introduces the reviewed entry labels used in our benchmark.

A common heuristic treats points far from the diagonal, which have high persistence, as candidate \textit{significant points} and points near the diagonal as candidate \textit{noise points}. This heuristic provides a reference for interpreting persistence-based labels.

\subsection{Confidence sets and two-means selection}

\begin{figure}[htbp]
	\centering
	\includegraphics[width=0.8\linewidth]{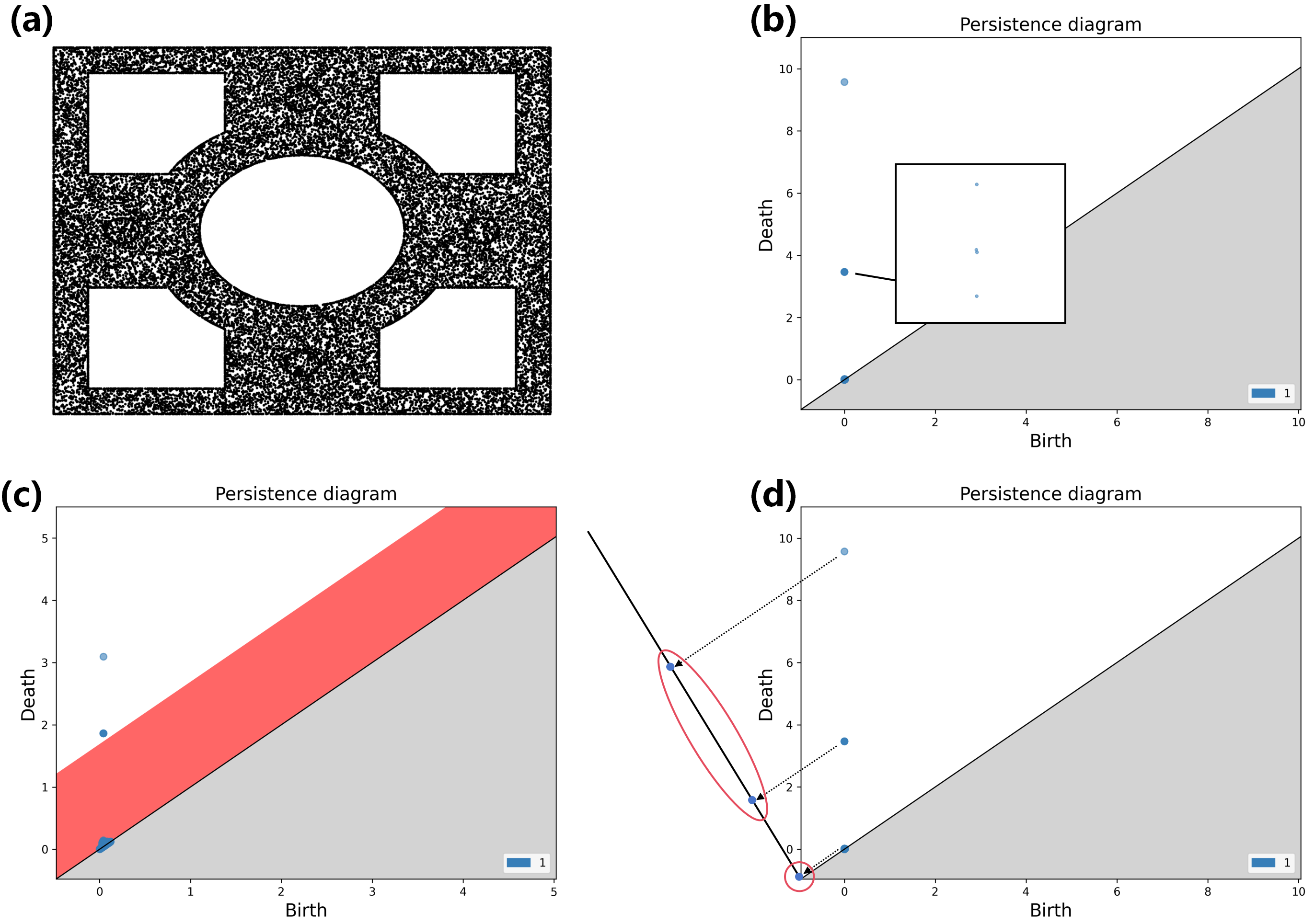}
    \caption{Illustration of persistence-based point labeling. (a) A planar point cloud with five loop features in the illustrative construction. (b) Its 1-PD. (c) The output of the confidence-set rule. (d) The output of the 2-means rule.}
    \Description{A point cloud with five holes appears beside its diagram, with an inset of the points near the diagonal. A red diagonal band marks the confidence-set rule in the third panel, and circles mark the two-means selections in the fourth.}
    \label{fig:CS_CLUS}
\end{figure}

Figure~\ref{fig:CS_CLUS} compares the confidence-set and 2-means rules. Panel (a) shows a planar point cloud whose illustrative construction contains five loops, and panel (b) shows its 1-PD, which contains five points with comparatively large persistence. In panel (c), the confidence-set rule places a statistically derived band around the diagonal at confidence level 0.5; entries inside the band are labeled as noise and entries outside it as significant~\citep{fasy2014confidence}. In panel (d), 2-means partitions the retained $(b_i,d_i)$ coordinates directly into two clusters. The cluster with larger mean persistence is labeled as significant.

\FloatBarrier

\section{Benchmark and Entry-Level Labels}\label{sec:datasets}

\subsection{Source geometries and observations}

The benchmark covers four categories of point-cloud data, (i) planar shapes; (ii) CAD models drawn from the ABC dataset
\footnote{\url{https://deep-geometry.github.io/abc-dataset}}
\citep{koch2019abc} and the Thingi10k dataset
\footnote{\url{https://github.com/thingi10k/thingi10k}}
\citep{zhou2016thingi10k}; (iii) triply periodic minimal surface (TPMS) structures
\footnote{\url{https://github.com/W-W-M/Morphology-Learning}}
\citep{wang2025simultaneous}; and (iv) zeolite crystal frameworks
\footnote{\url{http://america.iza-structure.org/IZA-SC/ftc_table.php}}
\citep{baerlocher2007atlas}. 
These categories range from controlled planar loop constructions to heterogeneous CAD surfaces, periodic multi-scale tunnels, and porous crystalline geometries.

Our dataset contains 4,990 PD--point-cloud--label records. The development pool provides 3,000 training and 500 validation observations, and the four test sets contain 250 planar, 500 CAD, 500 TPMS, and 240 zeolite observations. The partitions contain distinct sampled observations. Incomplete source identifiers limit verification of disjointness at the level of the underlying shapes. For each observation, we compute its alpha diagram with GUDHI and retain up to 100 entries. The model also receives up to 4,096 sampled points, with masks excluding padding. Appendix~\ref{app:benchmark} gives source examples and sampling details.

\subsection{Reviewed label construction}\label{sec:label_construction}

The label for each entry records whether it belongs to the reviewed set of source loops. We initialize that set with the $\beta_1$ longest intervals, where $\beta_1$ is the source loop count supplied by the geometric construction or topological metadata. One author then reviews the entries against the source geometry and may replace a late-born positive with an earlier-born candidate, preserving $\beta_1$ positives. The reviewed files are frozen before training. At inference, TUN receives the diagram and sampled point cloud.

The benchmark uses the source count $\beta_1$ to initialize each retained 1-PD. GUDHI stores squared radii $(\widetilde b_i,\widetilde d_i)=(b_i^2,d_i^2)$. The model equations use these stored squared-radius coordinates and omit the tildes. The label construction has three steps.
\begin{enumerate}
    \item \textbf{Compute $\beta_1$.} For planar shapes, the geometric construction directly supplies the number of loops. For 3D models, it supplies the relevant topological metadata. For a connected orientable surface, we use $\beta_1=2g$ when the surface is closed and $\beta_1=2g+q-1$ when it has $q>0$ boundary components.
    \item \textbf{Rank and assign.} We rank retained H1 entries by native squared-radius persistence ($\widetilde d_i-\widetilde b_i$), select the top-$\beta_1$ entries as the initial positive set, and assign $y_i=0$ to the remainder.
    \item \textbf{Early-birth review.} One author inspects each exported label file with its source geometry. If needed, the review swaps an initially positive late-born entry with an earlier-born candidate while preserving exactly $\beta_1$ positive labels. Sampling scales differ across source geometries, so this decision is instance-specific. The corrected files are frozen before model training.
\end{enumerate}
At inference, the model receives neither the geometric/topological count nor the manual correction history. It predicts the final label for each retained entry from the PD and available point-cloud context.

\subsection{Early-birth separation}\label{sec:label_theory}

Early birth has a sufficient-condition interpretation. For a compact tame source $X$ with positive reach $\tau$ and a sample $C$ with $d_H(C,X)\leq\varepsilon<\tau/4$, the $\beta_1$ classes inherited from $X$ have birth at most $\varepsilon$ and lifetime at least $\tau-2\varepsilon$. Other sampled classes born by $\varepsilon$ have lifetime at most $2\varepsilon$. Proposition~\ref{prop:early_birth} establishes this separation in radius units. Reach and sampling-error bounds are unmeasured for individual benchmark instances, so the practical labels use geometric review.

We first state the conditions under which an early-birth persistence rule identifies the one-dimensional classes inherited from an underlying shape. Let $X\subset\mathbb{R}^3$ be a compact tame set with positive reach $\tau$ and finite first Betti number $\beta_1$. Let $C$ be a finite point cloud such that $d_H(C,X)\leq\varepsilon<\tau/4$. The alpha complex of $C$ is homotopy equivalent, at each filtration scale, to the corresponding union of balls~\citep{edelsbrunner1995union}. The equality $\|d_X-d_C\|_\infty=d_H(X,C)$ then gives, by persistence-diagram stability, $d_B(D_1(X),D_1(C))\leq\varepsilon$~\citep{cohen2005stability,chazal2009sampling}.

\begin{proposition}[Early-birth separation]\label{prop:early_birth}
Under the conditions above, $D_1(C)$ contains $\beta_1$ classes satisfying
\begin{equation}\label{eq:inherited_bounds}
 b\leq\varepsilon,\qquad d\geq\tau-\varepsilon,
 \qquad d-b\geq\tau-2\varepsilon.
\end{equation}
Every other class in $D_1(C)$ with $b\leq\varepsilon$ has $d-b\leq2\varepsilon$. Consequently, the $\beta_1$ most persistent classes among the early-born entries are precisely the classes matched to the degree-one homology of $X$.
\end{proposition}

\begin{proof}
For $0\leq r<\tau$, nearest-point projection makes the offset $X^r$ homotopy equivalent to $X$. Thus the ideal offset persistence has $\beta_1$ classes born at zero that survive to at least $\tau$, with no additional degree-one births before $\tau$. A bottleneck matching of cost at most $\varepsilon$ maps these classes to entries satisfying Eq.~\eqref{eq:inherited_bounds}. Any remaining sampled entry with $b\leq\varepsilon$ cannot match an ideal class born at or after $\tau$ and must instead match the diagonal, which implies $d-b\leq2\varepsilon$. Since $\varepsilon<\tau/4$, we have $\tau-2\varepsilon>2\varepsilon$.
\end{proof}

The implementation uses GUDHI alpha-complex filtration values, which are squared radii~\citep{maria2014gudhi}. Writing $(\widetilde b,\widetilde d)=(b^2,d^2)$, an inherited entry satisfies $\widetilde d-\widetilde b\geq\tau^2-2\tau\varepsilon$, whereas an early-born diagonal-matched entry satisfies $\widetilde d-\widetilde b\leq8\varepsilon^2$. The same condition $\varepsilon<\tau/4$ makes the first bound strictly larger, supporting ranking by native squared-radius persistence among early-born entries.

For a source mesh $M$, suppose $d_H(M,X)\leq\delta_M$, the Poisson-disk sample covers $M$ within $\delta$, and each perturbation has norm at most $\nu$. Then $d_H(C,X)\leq\delta_M+\delta+\nu$. The reach and these error bounds are unmeasured on individual benchmark samples, so Proposition~\ref{prop:early_birth} motivates the review criterion under sufficient conditions. A loop born in an offset $X^r$ can persist even when absent from $X$ itself. Such a loop falls outside the source-loop target.

\subsection{Label audit}\label{sec:label_audit}

The review changes which entries are positive in 1,269 of the 4,990 diagrams, with 5,284 replacements in total. Among test diagrams, the change rate is highest on TPMS (48.2\%) and zeolites (41.3\%). At the reference positive count, lifetime ranking gives F1 values of $0.7990$ and $0.8837$ on these two categories. These disagreements define the interval-selection task that the learned models address.

We audit the released labels against the count-and-persistence initialization described in Section~\ref{sec:label_construction}. For each diagram, we reconstruct the initial positive set by taking the top $k$ entries under native squared-radius persistence, where $k$ is the positive count in the final label file. A diagram is marked as changed when this reconstructed set differs from the final positive set. The number of swaps is the size of the final positive set outside the reconstructed set. Table~\ref{tab:dataset_label_audit} records 5,284 swaps across 1,269 of the 4,990 diagrams.

\begin{table}[htbp]
    \caption{Dataset composition and label audit. Positive rate is computed over valid retained PD entries. Changed diagrams differ from the reconstructed top-$\beta_1$ persistence initialization; swaps count positive-set replacements. The released neural-model split contains 3,000 training and 500 validation records; SVM and MLP construct an independent 3,000/500 sample-level split for each seed.}
    \label{tab:dataset_label_audit}
    \centering
    \footnotesize
    \setlength{\tabcolsep}{4pt}
    \begin{tabular}{lrrrrr}
        \toprule
        \textbf{Split} & \textbf{Records} & \textbf{Valid entries} & \textbf{Positive rate} & \textbf{Changed diagrams} & \textbf{Swaps} \\
        \midrule
        Development pool & 3,500 & 347,605 & $9.06\%$ & 851 ($24.3\%$) & 3,555 \\
        Planar & 250 & 25,000 & $5.54\%$ & 50 ($20.0\%$) & 68 \\
        CAD & 500 & 50,000 & $3.64\%$ & 28 ($5.6\%$) & 30 \\
        TPMS & 500 & 50,000 & $10.77\%$ & 241 ($48.2\%$) & 1,083 \\
        Zeolite & 240 & 24,000 & $19.63\%$ & 99 ($41.3\%$) & 548 \\
        \midrule
        Total & 4,990 & 496,605 & $9.02\%$ & 1,269 ($25.4\%$) & 5,284 \\
        \bottomrule
    \end{tabular}
\end{table}

Review changes the identity of positive entries most often on TPMS (48.2\%) and Zeolite (41.3\%) among the four test categories. We can also use this audit to score lifetime ranking at the \emph{known count}. For every diagram, the ranking selects exactly as many entries as the reference labels. If a test category has $P$ positive entries and $S$ swaps, the ranking has $P-S$ true positives, $S$ false positives and $S$ false negatives. Its pooled point-level F1 is therefore $1-S/P$. This gives $0.9509$ on Planar ($S/P=68/1386$), $0.9835$ on CAD ($30/1821$), $0.7990$ on TPMS ($1083/5387$) and $0.8837$ on Zeolite ($548/4710$). This diagnostic uses the reference count at test time and measures disagreement in the choice of entries at that count.

\FloatBarrier

\section{TopologyUnderstandingNet}\label{sec:methods}

\begin{figure}[htbp]
    \centering
    \includegraphics[width=\linewidth]{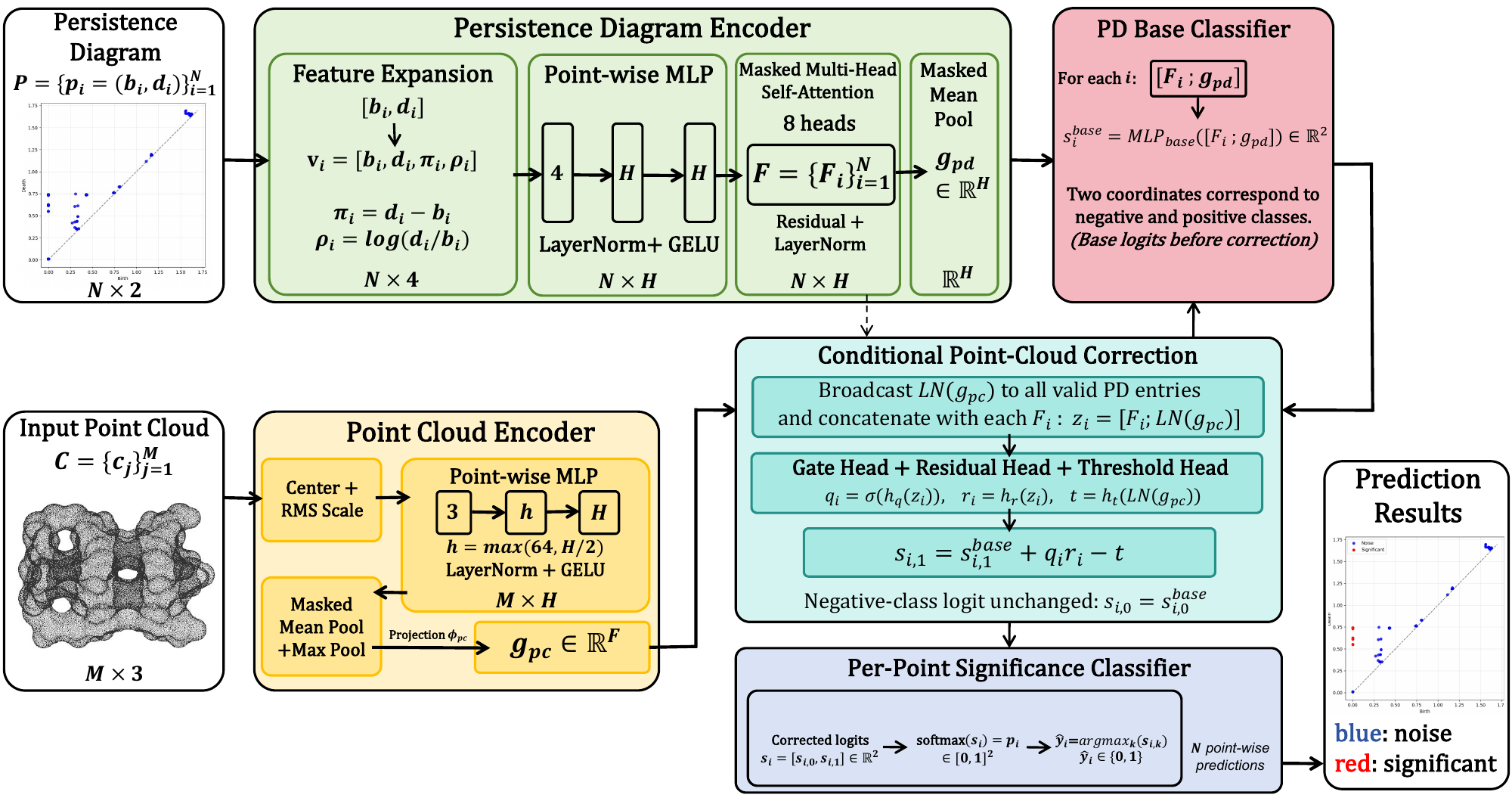}
    \caption{TUN architecture. Each retained birth--death pair is expanded to four descriptors and passed through a point-wise MLP, masked multi-head self-attention, residual normalization, and masked mean pooling. The resulting per-point features and global PD descriptor produce base two-class logits. In parallel, the source point cloud is centered, RMS-scaled, encoded point-wise, and summarized by masked mean and max pooling. Its projected descriptor drives a point-wise gated residual head and a sample-level threshold head that correct only the positive-class logit. The dashed arrow carries the per-entry PD features to the correction module. Equation~\eqref{eq:pd_features} gives the guarded log ratio used in the implementation.}
    \Description{Across the top, the diagram encoder feeds a base classifier. Below, a point-cloud encoder pools coordinates into a shared descriptor. The correction module combines this descriptor with diagram features and adjusts the positive logits before entry-wise classification.}
    \label{fig:architecture}
\end{figure}

TUN combines diagram context with a shared descriptor of the sampled geometry to predict a binary label for each retained entry. The diagram branch supplies base logits, and the point-cloud branch supplies a shared threshold shift and a candidate-specific residual correction. Figure~\ref{fig:architecture} shows this prediction path. The architecture can change both the number of positive entries and their ordering within a diagram.

\subsection{Diagram encoder and base scores}\label{sec:diagram_encoder}

Let $m_i^{\mathrm{pd}}\in\{0,1\}$ indicate whether entry $i$ is valid. After clipping the native squared-radius coordinates of $p_i=(b_i,d_i)$ to $[0,100]$, we form

\begin{equation}
    \pi_i=\max(0,d_i-b_i),\quad
    \rho_i=\log\frac{\max(d_i,\varepsilon)}{\max(b_i,\varepsilon)},\quad
    v_i=[b_i,d_i,\pi_i,\rho_i]\in\mathbb{R}^{4},
    \label{eq:pd_features}
\end{equation}

where $\varepsilon=10^{-6}$. A shared MLP and masked self-attention turn these descriptors into contextualized tokens $F_i$. Their masked mean gives a permutation-invariant diagram descriptor $g_{\mathrm{pd}}$. A two-class head combines the token and global descriptor,

\begin{equation}
    s_i^{\mathrm{base}}=
    \operatorname{MLP}_{\mathrm{base}}\left([F_i;g_{\mathrm{pd}}]\right)
    \in\mathbb{R}^{2},
    \label{eq:pd_base_logits}
\end{equation}

with coordinates corresponding to negative and positive logits. This is the prediction path of the PD-Encoder only initialization.

Equation~\eqref{eq:pd_features} defines the four descriptors, with $\varepsilon=10^{-6}$ keeping the logarithm finite at zero. We replace non-finite derived values with finite constants. A shared point-wise MLP maps each $v_i$ through $\operatorname{Linear}(4,H)$, LayerNorm, GELU, dropout, $\operatorname{Linear}(H,H)$ and a second LayerNorm. The resulting $N_{\mathrm{pd}}\times H$ token matrix passes to an eight-head masked self-attention layer in PyTorch. The residual sum of the input tokens and the attention output is normalized by LayerNorm. Padding tokens are zeroed before attention, excluded as attention keys and values, and masked again after normalization. The encoder produces contextualized features $F_i$ for valid PD entries and a permutation-invariant global descriptor
\begin{equation}
    g_{\mathrm{pd}}=
    \frac{\sum_i m_i^{\mathrm{pd}}F_i}{\max\left(1,\sum_i m_i^{\mathrm{pd}}\right)}.
    \label{eq:pd_pooling}
\end{equation}

The base classifier in Eq.~\eqref{eq:pd_base_logits} applies $\operatorname{Linear}(2H,F)$, LayerNorm, GELU, dropout, and $\operatorname{Linear}(F,2)$ to $[F_i;g_{\mathrm{pd}}]$.

\subsection{Point-cloud encoder}\label{sec:pc_encoder}

A masked PointNet-style encoder~\citep{qi2017pointnet} centers the point cloud, normalizes its RMS radius, and pools point-wise features by their mean and maximum. A projection produces the descriptor $g_{\mathrm{pc}}$.

To incorporate geometric context, we use a masked PointNet-style encoder inspired by~\citep{qi2017pointnet}. Two-dimensional inputs are embedded in $\mathbb{R}^3$ with a zero $z$ coordinate. Let $m_j^{\mathrm{pc}}$ denote the validity mask for point $c_j\in\mathbb{R}^{3}$. We center the valid points using
\begin{equation}
    \bar c=\frac{\sum_jm_j^{\mathrm{pc}}c_j}{\max(1,\sum_jm_j^{\mathrm{pc}})},\quad
    a=\left(\frac{\sum_jm_j^{\mathrm{pc}}\lVert c_j-\bar c\rVert_2^2}
    {\max(1,\sum_jm_j^{\mathrm{pc}})}+10^{-8}\right)^{1/2}.
    \label{eq:pc_normalization}
\end{equation}
We then scale the points as $\widetilde c_j=m_j^{\mathrm{pc}}(c_j-\bar c)/\max(a,10^{-4})$. A shared point-wise MLP applies $\operatorname{Linear}(3,h)$ and $\operatorname{Linear}(h,H)$, each followed by LayerNorm and GELU, with $h=\max(64,H/2)$. If $h_j^{\mathrm{pc}}\in\mathbb{R}^{H}$ denotes the resulting feature, the two masked pooling outputs are
\begin{equation}
    u_{\mathrm{mean}}=\frac{\sum_jm_j^{\mathrm{pc}}h_j^{\mathrm{pc}}}
    {\max(1,\sum_jm_j^{\mathrm{pc}})},\quad
    u_{\mathrm{max}}=\operatorname*{max}_{j:m_j^{\mathrm{pc}}=1}h_j^{\mathrm{pc}},\quad
    g_{\mathrm{pc}}=\phi_{\mathrm{pc}}([u_{\mathrm{mean}};u_{\mathrm{max}}])
    \in\mathbb{R}^{F}.
    \label{eq:pc_pooling}
\end{equation}
Here $\phi_{\mathrm{pc}}$ applies $\operatorname{Linear}(2H,F)$, LayerNorm, GELU, and dropout. Masking excludes padded points and their placeholder coordinates from the descriptor.

We also pass this descriptor to a point-cloud count head, $\widehat c_{\mathrm{pc}}=\operatorname{softplus}(u_{\mathrm{pc}}(g_{\mathrm{pc}}))$. This head supplies an auxiliary training signal and is not an input to the per-point classifier.

\subsection{Conditional logit correction}\label{sec:conditional_correction}

For each PD entry, we concatenate the normalized descriptor with its token as $z_i=[F_i;\operatorname{LN}(g_{\mathrm{pc}})]$. The conditional heads produce

\begin{equation}
    r_i=h_r(z_i),\quad q_i=\sigma(h_q(z_i))\in(0,1),\quad
    t=h_t(\operatorname{LN}(g_{\mathrm{pc}})),
    \label{eq:conditional_heads}
\end{equation}

where $r_i$ is a residual, $q_i$ is its gate, and $t$ is shared across the diagram. They correct the positive-class logit according to

\begin{equation}
 s_{i,1}=s^{\mathrm{base}}_{i,1}+q_i r_i-t.
 \label{eq:gated_residual}
\end{equation}

The negative-class logit comes from the PD branch. The shared threshold $t$ shifts the positive log-odds equally for all entries, while $q_i r_i$ can change their relative order. Softmax yields class probabilities, and the larger logit determines the predicted label. Each correction uses the same cloud descriptor together with its own diagram token.

For the correction in Eq.~\eqref{eq:gated_residual}, we normalize $g_{\mathrm{pc}}$ with LayerNorm and broadcast it to the valid PD tokens to form $z_i$.

The residual and gate heads in Eq.~\eqref{eq:conditional_heads} produce one value per valid PD entry, whereas the threshold head produces one scalar shared by all entries in a sample. The residual adapter maps $z_i$ through $\operatorname{Linear}(H+F,F)$, LayerNorm, GELU, dropout, and $\operatorname{Linear}(F,1)$. The gate maps $z_i$ through $\operatorname{Linear}(H+F,F/2)$, GELU, and $\operatorname{Linear}(F/2,1)$ before the sigmoid. The threshold head applies $\operatorname{Linear}(F,F/2)$, GELU, and $\operatorname{Linear}(F/2,1)$ to the normalized cloud descriptor.

\subsection{Training objective}\label{sec:loss}

Training combines a weighted focal classification loss~\citep{lin2017focal}, two count losses, and a residual penalty,

\begin{equation}
    \mathcal{L}_{\mathrm{total}}
    =\mathcal{L}_{\mathrm{focal}}
    +\lambda_c\mathcal{L}_{\mathrm{count}}
    +\lambda_{\mathrm{pc}}\mathcal{L}_{\mathrm{pc-count}}
    +\lambda_r\mathcal{L}_{\mathrm{res}}.
    \label{eq:total_loss}
\end{equation}

The first count term supervises the sum of positive probabilities, and the second supervises a count head on $g_{\mathrm{pc}}$. Both use smooth-$L_1$ loss on log-counts. The residual penalty averages $r_i^2$ over valid entries. We use weights $(\lambda_c,\lambda_{\mathrm{pc}},\lambda_r)=(0.1,0.05,10^{-4})$. PD-Encoder only uses the same focal term and expected-count loss.

We adopt a weighted focal loss~\citep{lin2017focal} because significant points are rare relative to noise points. For each PD point $i$, let $\mathbf{s}_i\in\mathbb{R}^2$ denote the two-class logits, let $\mathbf{p}_i=\operatorname{softmax}(\mathbf{s}_i)$ denote the class probabilities, and let $y_i\in\{0,1\}$ be the ground-truth label. We write $p_{t,i}=\mathbf{p}_i[y_i]$ for the probability assigned to the true class and define the weighted cross-entropy value $e_i=w_{y_i}(-\log p_{t,i})$. We then apply the focal modulator to this weighted value:
$ \ell_i = \alpha\, \big(1-\exp(-e_i)\big)^{\gamma}\, e_i$.
Here $w_{y_i}\in\mathbb{R}_+$ is the class weight of label $y_i$ (with $w_1>w_0$), $\alpha>0$ is a global scaling coefficient, and $\gamma\ge 0$ down-weights easy examples. Because PDs are padded to a fixed length, we compute the loss only over valid points using a mask $M_i\in\{0,1\}$. The focal loss averages these terms over the valid entries,
\[ \mathcal{L}_{\mathrm{focal}} = \frac{1}{\sum_i M_i} \sum_i M_i\, \ell_i. \]

We supplement the point loss with two count terms and a residual regularizer. Let $\hat c$ be the sum of predicted positive probabilities over valid entries and let $c$ be the target positive count:
\begin{equation}
    \widehat c=\sum_iM_i p_{i,1},
    \qquad c=\sum_iM_i y_i.
    \label{eq:expected_count}
\end{equation}
The point-cloud count head gives a second estimate $\hat c_{\mathrm{pc}}$. Let $\mathcal{L}_{\mathrm{res}}$ denote the mean squared residual output over valid entries. The terms enter the objective in Eq.~\eqref{eq:total_loss}.
Both count terms use smooth-$L_1$ loss on log-counts. For the PD prediction, the loss compares $\log(1+\hat c)$ with $\log(1+c)$; for the point-cloud head, it compares $\log(1+\hat c_{\mathrm{pc}})$ with $\log(1+c)$. The residual penalty is applied to $r_i$ only on valid entries. We use $\lambda_c=0.1$, $\lambda_{\mathrm{pc}}=0.05$, and $\lambda_r=10^{-4}$ for the full TUN model. The matched PD-Encoder only model uses the PD count term but has no point-cloud count or residual branch. All point-level terms exclude padded entries through the valid-entry mask.

\subsection{Training settings and computational cost}\label{sec:training}

Each TUN run starts from its same-seed PD-only checkpoint. Residual and threshold outputs start at zero, the gate bias is $-2$, and transferred parameters are frozen for the first ten epochs. We use AdamW with cosine annealing and select the checkpoint with minimum validation loss. The ten-run evaluation shares one configuration across all four categories.

Unless otherwise specified, we use $H=256$, fusion dimension $F=256$, intermediate point-cloud width $h=128$, eight attention heads, and dropout $0.2$. The weighted focal loss uses $(\alpha,\gamma)=(1.0,2.0)$ and class weights $w_0=1.0$, $w_1=2.0$.

We train with AdamW and cosine annealing and clip gradients to a global norm of 1.0. The batch size is 16 and the maximum number of epochs is 200. Each persistence diagram is retained at $N_{\text{pd}}=100$ entries, and each point cloud is subsampled or padded to $N_{\text{pc}}=4,096$ entries. Valid lengths exclude padding from the objective and metrics, and the point-cloud encoder applies per-cloud centering and RMS scaling. Experiments ran on one NVIDIA GeForce RTX 4090 GPU (24 GB), a 16-core Intel Xeon Platinum 8352V CPU at 2.10 GHz, and 120 GB of system RAM.

The PD-Encoder only configuration uses a learning rate of $0.001$. The final full TUN model uses a learning rate of $0.0005$ for newly added context parameters and $0.0001$ for transferred PD parameters, with weight decay $0.01$ and a 10-epoch warm-start freeze. Training stops early when the validation loss fails to improve by $10^{-4}$ for 10 epochs. Each configuration is trained in ten independent runs, and the checkpoint with minimum validation loss is used for testing. One configuration is shared across the four categories; test-set metrics are not used for checkpoint or hyperparameter selection.

Let $N$ denote the retained number of PD entries, let $M$ denote the retained number of point-cloud samples, and let $H$ denote the hidden width. PD self-attention costs $O(N^2H)$ time and $O(N^2)$ attention memory. The point-cloud encoder uses shared point-wise layers and masked pooling, which cost $O(MH^2)$ time and $O(MH)$ activation memory for fixed-width layers. The conditional correction is point-wise and costs $O(NH^2)$. With the implemented caps $N=100$ and $M=4{,}096$, the point-cloud branch dominates model-side activation storage. Alpha-complex and persistence-diagram construction remain separate preprocessing costs.

\FloatBarrier

\section{Classification of Diagram Entries}\label{sec:results}

\subsection{Evaluation protocol}

We evaluate agreement with the reviewed entry-level labels using point-level F1. The comparisons include 2-means and a confidence-set rule, point-wise SVM and MLP classifiers, DeepSets and Set Transformer with entry-wise heads, the PD-Encoder only initialization, and full TUN. The neural set models use the common 3,000/500 training/validation split. SVM and MLP construct a 3,000/500 split from the same development pool at each seed. All methods use the same test observations. Table~\ref{tab:main_results} reports means and sample standard deviations over ten runs for learned models, with deterministic rules evaluated once.

\subsection{Comparison methods}\label{sec:baselines}

On each test category we evaluate 2-means clustering and a confidence-set (CS) rule at confidence level $0.5$ once each. Both operate in radius units. We transform the native squared-radius coordinates $(\widetilde b_i,\widetilde d_i)$ to $(b_i,d_i)=(\sqrt{\widetilde b_i},\sqrt{\widetilde d_i})$ while preserving entry order. For 2-means, we apply $k$-means with $k=2$ and ten initializations. The cluster with larger mean radius persistence is positive. For CS(0.5), each point cloud of size $n$ is resampled 500 times with replacement using subsamples of size $\lfloor n/\log n\rfloor$. In each repetition we measure the maximum nearest-neighbor distance from the held-out points to the subsample. If $q_{0.5}$ is the median of these distances, an entry is positive when $d_i-b_i>4q_{0.5}$. The resampling seed starts at zero and is incremented by sample index.

All supervised comparison methods predict one binary label for each retained PD entry. The neural set models and TUN variants use the released fixed sample-level split of the 3,500-record development pool into 3,000 training and 500 validation records for all ten initializations. For SVM and the point-wise MLP, each seed independently constructs a 3,000/500 sample-level split from the same development pool. SVM~\citep{hearst1998support} uses scikit-learn's RBF classifier with $C=1$, $\gamma=\mathtt{scale}$, and balanced class weights on the clipped but otherwise unstandardized coordinates. The two-layer MLP~\citep{murtagh1991multilayer} uses 64 hidden units per layer, ReLU, BatchNorm, dropout $0.3$, AdamW with learning rate $0.001$, batch size 16, and inverse-frequency weighted cross-entropy. Its coordinates are standardized using statistics fitted only on that seed's training split, and training stops by validation loss with patience 20 or after 200 epochs. Both SVM and MLP receive only $(b_i,d_i)\in\mathbb{R}^2$.

DeepSets~\citep{zaheer2017deep} and Set Transformer~\citep{lee2019set} are PD-only set baselines adapted with point-wise heads. Both use width 256, dropout $0.2$, and the same four derived PD descriptors as TUN. Their global set descriptor also drives an auxiliary log-count regression head, trained with weight $0.1$. DeepSets combines shared token layers with masked-mean aggregation. Set Transformer uses two eight-head self-attention blocks and multi-head-attention pooling. Both preserve point-level permutation equivariance while using permutation-invariant global context. These two set baselines use the same valid-entry masks, ten-run protocol, validation-loss checkpoint rule, and point-wise metrics as TUN; the point-wise MLP instead operates directly on the unpadded retained entries.

\subsection{Classification accuracy and error composition}

TUN attains the highest F1 on Planar, TPMS, and Zeolite, with DeepSets higher on CAD by $0.0010$ (Table~\ref{tab:main_results}). Its unweighted four-category mean F1 is $0.9965$, compared with $0.9933$ for DeepSets, $0.9926$ for Set Transformer, and $0.9927$ for PD-Encoder only. The sample standard deviation of the seed-wise four-category mean is $0.0010$ for TUN, compared with $0.0032$, $0.0032$, and $0.0036$ for these three PD-only models.

\begin{table}[htbp]
    \caption{Point-level F1 by test category. Deterministic rules are evaluated once. Learning methods report mean $\pm$ sample standard deviation over ten runs. CS uses confidence level $0.5$ and $B=500$ resamples. PD-Encoder only supplies the initialization for full TUN. Bold marks the best value in each column.}
    \label{tab:main_results}
    \centering
    \small
    \begin{tabular}{lcccc}
        \toprule
        \textbf{Method} & \textbf{Planar} & \textbf{CAD} & \textbf{TPMS} & \textbf{Zeolite} \\
        \midrule
        2-means & $0.3071$ & $0.1627$ & $0.3224$ & $0.5357$ \\
        CS ($0.5$) & $0.2525$ & $0.1263$ & $0.0000$ & $0.0000$ \\
        MLP & $0.5143\!\pm\!0.1207$ & $0.2693\!\pm\!0.0923$ & $0.3243\!\pm\!0.1502$ & $0.0743\!\pm\!0.1017$ \\
        SVM & $0.9296\!\pm\!0.0016$ & $0.8701\!\pm\!0.0024$ & $0.9653\!\pm\!0.0007$ & $0.8880\!\pm\!0.0015$ \\
        DeepSets & $0.9874\!\pm\!0.0056$ & $\mathbf{0.9945\!\pm\!0.0031}$ & $0.9926\!\pm\!0.0049$ & $0.9987\!\pm\!0.0007$ \\
        Set Transformer & $0.9849\!\pm\!0.0055$ & $0.9919\!\pm\!0.0102$ & $0.9951\!\pm\!0.0031$ & $0.9987\!\pm\!0.0010$ \\
        \midrule
        PD-Encoder only & $0.9878\!\pm\!0.0032$ & $0.9883\!\pm\!0.0113$ & $0.9965\!\pm\!0.0027$ & $0.9981\!\pm\!0.0017$ \\
        \textbf{TUN} & $\mathbf{0.9941\!\pm\!0.0019}$ & $0.9935\!\pm\!0.0029$ & $\mathbf{0.9986\!\pm\!0.0014}$ & $\mathbf{0.9996\!\pm\!0.0007}$ \\
        \bottomrule
    \end{tabular}
\end{table}

The F1 gain over the PD-only initialization comes primarily from fewer false positives. TUN improves four-category mean F1 in all ten paired runs, with an average gain of $0.0038$. Per run, false positives fall from $113.7$ to $25.0$, while false negatives rise from $20.0$ to $33.7$. On CAD, TUN also produces fewer false positives than DeepSets ($8.1$ versus $15.5$), whose higher recall gives the higher F1. Appendix~\ref{app:classification} gives the confusion counts. Section~\ref{sec:short_intervals} gives a short-interval example.

\subsection{Ranking of retained entries}\label{sec:entry_ranking}

Table~\ref{tab:set_ap} evaluates the ranking of valid entries pooled across diagrams within each test category, using the predicted probabilities for the positive class. Full TUN's seed-averaged AP exceeds that of its PD-only initialization in each category. DeepSets ranks highest on Planar and Set Transformer on CAD and Zeolite. Because the rankings pool diagrams, a diagram-specific score shift can also affect AP.

\begin{table}[htbp]
    \centering
    \caption{Point-level average precision (AP). We pool valid entries within each test category, compute AP for each seed, and report the mean $\pm$ sample standard deviation over ten seeds. Bold marks the highest mean in each column. Extra digits distinguish the near-ceiling scores on Zeolite.}
    \label{tab:set_ap}
    \footnotesize
    \setlength{\tabcolsep}{3pt}
    \begin{tabular}{lcccc}
        \toprule
        \textbf{Method} & \textbf{Planar} & \textbf{CAD} & \textbf{TPMS} & \textbf{Zeolite} \\
        \midrule
        DeepSets & $\mathbf{0.99994\!\pm\!0.00005}$ & $0.99969\!\pm\!0.00064$ & $0.99966\!\pm\!0.00023$ & $0.9999973\!\pm\!0.0000027$ \\
        Set Transformer & $0.99956\!\pm\!0.00032$ & $\mathbf{0.99989\!\pm\!0.00030}$ & $0.99989\!\pm\!0.00014$ & $\mathbf{0.9999996\!\pm\!0.0000006}$ \\
        PD-Encoder only & $0.99846\!\pm\!0.00155$ & $0.99761\!\pm\!0.00684$ & $0.99988\!\pm\!0.00014$ & $0.9999950\!\pm\!0.0000087$ \\
        TUN & $0.99916\!\pm\!0.00072$ & $0.99974\!\pm\!0.00023$ & $\mathbf{0.99994\!\pm\!0.00008}$ & $0.9999994\!\pm\!0.0000009$ \\
        \bottomrule
    \end{tabular}
\end{table}

\subsection{Selection at the reviewed count}\label{sec:selection_controls}

Supplying the reviewed positive count isolates the choice of entries at that count. On the 418 test diagrams changed by geometric review, TUN and PD-only each recover the exact positive set in 4,179 of 4,180 diagram--checkpoint evaluations. Both score orderings closely reproduce the reviewed replacements. This diagnostic explains why the complete positive-set decision, including its count, matters when interpreting the F1 gains.

Let $Y$ be the reviewed set and $L$ the $k=|Y|$ longest intervals in squared-radius units. On diagrams with $L\ne Y$, each model selects its $k$ highest positive probabilities. We count selected members of $Y\setminus L$ and excluded members of $L\setminus Y$. The saved PD rows and labels agree across models and reproduce the released exchange counts. No ties straddle the cutoff on these diagrams. This diagnostic supplies the reference count at evaluation. At this count, TUN and PD-only each recover the exact positive set in 4,179 of 4,180 evaluations of the 418 test diagrams with exchanges.

\begin{table}[htbp]
    \centering
    \caption{Known-count selection on changed test diagrams. Model columns give mean exchanged-in entries recovered per checkpoint. Exchanged-out entries correctly excluded have the same counts in every category and checkpoint.}
    \label{tab:exchange_selection}
    \small
    \begin{tabular}{lrrrrr}
        \toprule
        & Changed diagrams & Positives & Exchanges & TUN & PD-only \\
        \midrule
        Planar & 50 & 337 & 68 & 68.0 & 68.0 \\
        CAD & 28 & 138 & 30 & 30.0 & 30.0 \\
        TPMS & 241 & 3,365 & 1,083 & 1,082.7 & 1,082.9 \\
        Zeolite & 99 & 2,526 & 548 & 548.0 & 548.0 \\
        \bottomrule
    \end{tabular}
\end{table}

All errors in Table~\ref{tab:exchange_selection} occur on TPMS. TUN misses three exchanged-in entries on one diagram at seed 33. PD-only misses one on a different diagram at seed 1. Both score orderings closely reproduce the annotator's decisions.

\subsection{Point-cloud pairing}\label{sec:cloud_pairing}

Point-cloud permutation probes the contribution of sample-specific pairing. Permuting clouds within each same-category batch changes mean F1 by at most $0.00094$ in absolute value on Planar, CAD, and TPMS. Most of the gain over PD-only persists under this permutation. Full TUN also adds point-cloud count supervision and continued training to its PD-only initialization, so the overall comparison includes these changes.

An archived control permutes clouds and valid lengths within each same-category batch (up to 16 observations), keeping PDs and labels fixed. Its paired reference permutes an unused auxiliary input. On the three splits in Table~\ref{tab:cloud_permutation}, PD rows and labels align across controls, and paired probabilities and predictions exactly match final TUN outputs at every seed. The archived Zeolite control predates the change from 4,752 to 4,710 test positives.

\begin{table}[htbp]
    \centering
    \caption{Mean point-level F1 with the paired or batch-permuted point cloud. Differences are computed within checkpoint before averaging. SD describes the variation in paired differences across ten checkpoints.}
    \label{tab:cloud_permutation}
    \small
    \begin{tabular}{lrrrr}
        \toprule
        & Paired & Permuted & Paired minus permuted & SD \\
        \midrule
        Planar & 0.99410 & 0.99317 & $+0.00094$ & 0.00074 \\
        CAD    & 0.99350 & 0.99341 & $+0.00009$ & 0.00551 \\
        TPMS   & 0.99862 & 0.99878 & $-0.00016$ & 0.00050 \\
        \bottomrule
    \end{tabular}
\end{table}

Most of the mean F1 advantage over PD-only persists under within-category permutation. Some clouds can retain their original pairing. On CAD the variation across checkpoints exceeds the mean pairing effect. The paired-minus-permuted mean F1 difference is at most $0.00094$ in absolute value across these three categories. The comparison between full TUN and its PD-only initialization also includes point-cloud count supervision and continued training.

\subsection{Short-interval example}\label{sec:short_intervals}

\begin{figure}[htbp]
    \centering
    \includegraphics[width=0.8\linewidth]{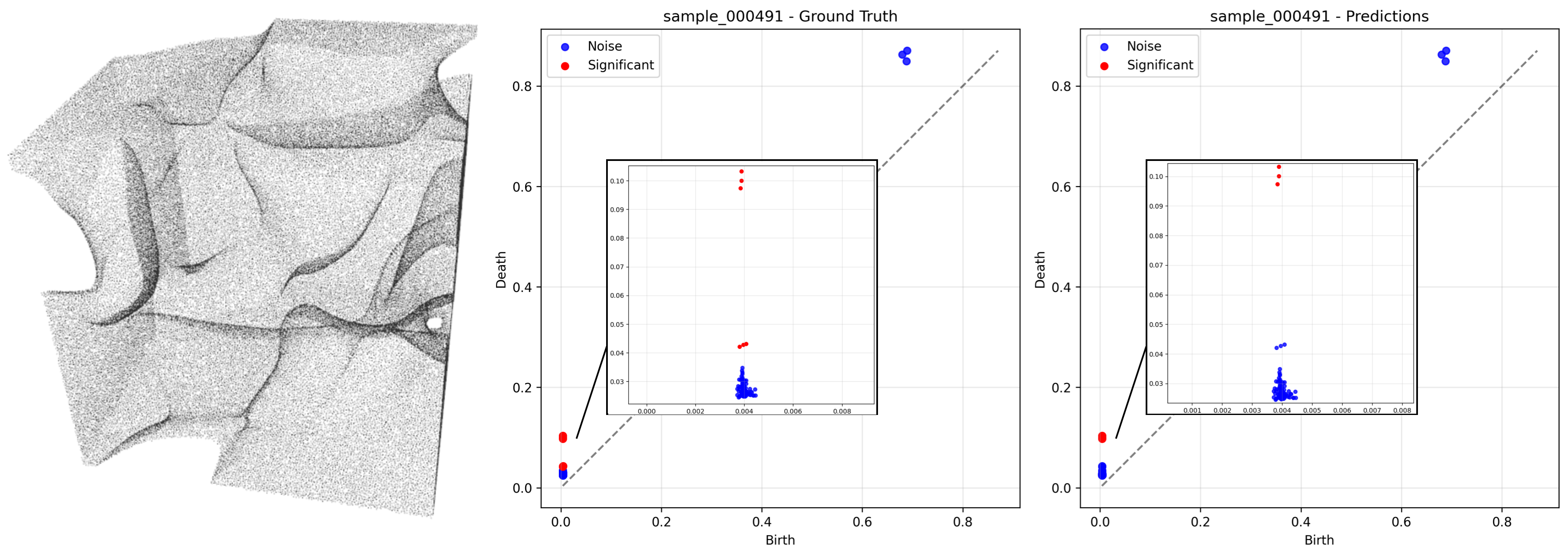}
    \caption{A TPMS test diagram in which TUN misses three reviewed positives. The three short intervals are labeled as small-scale source loops (red in the reference labels, blue in the predictions).}
    \Description{A point-cloud rendering sits beside reference and prediction diagrams. Insets enlarge the short intervals near death 0.04. Three entries that appear red in the reference inset are blue in the prediction inset.}
    \label{fig:limit}
\end{figure}

Figure~\ref{fig:limit} shows a TPMS test diagram in which TUN misses three short reviewed positives. Reference labels reflect one author's review of the ranked entries against the source geometry. An independent annotation would assess how consistently the exchanged entries are selected. TUN represents the cloud through a single shared vector, motivating the candidate-local geometric features discussed in Section~\ref{sec:conclusion}.

\section{CAD Source-Identity Retrieval}\label{sec:retrieval}

Retrieval tests whether selected intervals help identify the source of an independently sampled point cloud. Each query has exactly one correct clean mesh in the gallery, making source identity the evaluation target.

\subsection{Gallery and query construction}\label{sec:retrieval_protocol}

We use 500 Thingi10K meshes selected by geometric criteria, assigning 100 source identities to protocol validation and 400 to final evaluation. The final gallery contains 400 clean, loop-bearing models. Each query is resampled independently from its source and receives one of nine corruptions covering reduced density, jitter, non-uniform sampling, added outliers, or partial observation.

\paragraph{Dataset and fixed split.}
The retrieval collection is selected from Thingi10K by geometry criteria alone. Auditing merges coincident vertices and requires finite coordinates, a non-degenerate triangulation, manifold edges, watertightness, and one connected component. All 500 retained meshes pass these checks and have mesh-derived first Betti numbers below 100. A fixed source-identity manifest assigns 100 meshes to protocol validation and 400 to final evaluation. The full-gallery experiment uses the 400 final-evaluation identities and therefore measures retrieval among loop-bearing CAD models.

\paragraph{Sampling, corruptions, and persistence diagrams.}
For a clean gallery item, 8,192 surface points are generated by area-weighted oversampling followed by farthest-point sampling. Each query is independently resampled from its source mesh and receives exactly one of the following transformations: \emph{density} at 50\% or 25\% of the nominal points; \emph{jitter} with Gaussian coordinate noise whose standard deviation is $0.005$ or $0.01$ times the source bounding-box diagonal; \emph{non-uniform} sampling with triangle probabilities proportional to squared triangle area; \emph{outliers} at 1\% or 5\% of the sampled points, drawn from a box padded by 5\% of the diagonal; or \emph{partial} observation with 20\% or 30\% of the points removed by a random-direction half-space quantile. STL coordinate units are preserved before persistence computation.

For each clean or corrupted observation, we use GUDHI to construct the alpha filtration on its dense point set and compute $H_1$ persistence over $\mathbb{Z}_2$. Filtration values are squared radii. After removing infinite intervals, we retain up to 100 entries by persistence. TUN also receives a deterministic subsample of at most 4,096 points from the same observation.

\subsection{Selection, distance, and evaluation}\label{sec:retrieval_evaluation}

For each query and each clean gallery item, TUN retains the entries it classifies as positive. Top-persistence selects the same number of longest intervals, and Random-matched averages five uniform subsets of that size. Count-only compares the absolute difference between the two selected counts. All diagram-valued selectors use the same finite-diagonal assignment distance in native squared-radius coordinates. The primary protocol searches all 400 clean meshes. A secondary protocol restricts each query's gallery to meshes with the same mesh-derived $\beta_1$. We report R@1, R@5, and mAP, which equals mean reciprocal rank in this single-relevant-item task. Conditions are averaged within checkpoint before summarizing the ten checkpoints.

At each seed, full TUN retains the entries it classifies as positive. Their number $k_{\mathrm{TUN}}$ is specific to each observation, including each clean gallery item. Top-persistence retains its $k_{\mathrm{TUN}}$ longest intervals. Random-matched draws five uniform subsets of the same size and averages their results. Count-only compares two observations by $|k_q-k_g|$.

We use a finite-diagonal augmented assignment distance for all diagram-valued selectors. An off-diagonal match between $x,y\in\mathbb{R}^2$ incurs $\lVert x-y\rVert_2^2$, an unmatched interval $(b,d)$ incurs $((d-b)/2)^2$, and ranking uses the square root of the minimum total assignment cost. The distance remains finite when one selected diagram is empty. The primary full-gallery protocol compares against all 400 clean final-evaluation models. The secondary protocol restricts the gallery to models with the same mesh-derived $\beta_1$ as the query, so its gallery size varies across queries. Both protocols use direct distance ranking without training a retrieval model.

Let $r_q$ be the rank of the unique clean gallery model with the same source identity as query $q$. For $Q$ queries,
\begin{equation}
    \mathrm{R@}K=\frac{1}{Q}\sum_{q=1}^{Q}\mathbb{I}[r_q\le K],
    \qquad
    \mathrm{mAP}=\frac{1}{Q}\sum_{q=1}^{Q}\frac{1}{r_q}.
\end{equation}
In this single-relevant-item task, mAP equals mean reciprocal rank. Each gallery protocol contributes $400\times9=3{,}600$ query cases per checkpoint. We first average the nine equally sized corruption conditions within each checkpoint and protocol, then report the mean and sample standard deviation across ten checkpoints. Random-matched is first averaged across its five subset draws.

\subsection{Retrieval gains under added outliers}

Across all nine corruptions, full-gallery mAP is $0.3540$ for TUN and $0.3215$ for lifetime ranking at the same per-observation counts (Table~\ref{tab:cad_retrieval_protocols}). The largest gains occur under added outliers. At 1\% outliers, TUN retrieves the correct mesh first in 1,223 of 4,000 query--checkpoint pairs, compared with 318 for lifetime ranking. The paired mAP difference is $+0.2331$ with a 95\% mesh-bootstrap interval of $[0.1978,0.2696]$. At 5\% outliers, the difference is $+0.0958$ with interval $[0.0721,0.1209]$. These intervals resample source identities while retaining their checkpoints and conditions, conditional on the observed corruptions and fixed gallery.

\begin{table}[htbp]
    \caption{Aggregate CAD retrieval by gallery protocol. Values are means and sample standard deviations over ten seeds. Random-matched is first averaged over five subset draws. Bold marks the largest displayed mean for each protocol and metric.}
    \label{tab:cad_retrieval_protocols}
    \centering
    \footnotesize
    \setlength{\tabcolsep}{6pt}
    \begin{tabular}{llccc}
        \toprule
        \textbf{Gallery protocol} & \textbf{Selection} & \textbf{R@1} & \textbf{R@5} & \textbf{mAP} \\
        \midrule
        Full gallery & Count-only & $0.0180\!\pm\!0.0019$ & $0.0763\!\pm\!0.0056$ & $0.0624\!\pm\!0.0026$ \\
        & Random-matched & $0.0424\!\pm\!0.0024$ & $0.1178\!\pm\!0.0062$ & $0.0885\!\pm\!0.0040$ \\
        & Top-persistence@$k_{\mathrm{TUN}}$ & $0.2618\!\pm\!0.0173$ & $0.3750\!\pm\!0.0226$ & $0.3215\!\pm\!0.0197$ \\
        & TUN & $\mathbf{0.2934\!\pm\!0.0222}$ & $\mathbf{0.4094\!\pm\!0.0267}$ & $\mathbf{0.3540\!\pm\!0.0240}$ \\
        \addlinespace
        Same-$\beta_1$ gallery & Count-only & $0.1384\!\pm\!0.0053$ & $0.3882\!\pm\!0.0059$ & $0.2658\!\pm\!0.0045$ \\
        & Random-matched & $0.1637\!\pm\!0.0066$ & $0.3861\!\pm\!0.0077$ & $0.2796\!\pm\!0.0069$ \\
        & Top-persistence@$k_{\mathrm{TUN}}$ & $0.4023\!\pm\!0.0193$ & $0.6163\!\pm\!0.0216$ & $0.5058\!\pm\!0.0190$ \\
        & TUN & $\mathbf{0.4345\!\pm\!0.0225}$ & $\mathbf{0.6438\!\pm\!0.0241}$ & $\mathbf{0.5349\!\pm\!0.0219}$ \\
        \bottomrule
    \end{tabular}
\end{table}

Lifetime ranking has a $0.0052$ higher mean mAP across the other seven conditions. The nine-condition aggregate therefore reflects the larger gains near first place under outliers. Count-only and Random-matched obtain lower aggregate mAP in both gallery protocols. Appendix~\ref{app:retrieval_results} reports each corruption. Section~\ref{sec:paired_ranks} reports paired rank changes, uncertainty, and the treatment of equal distances.

\subsection{Paired rank changes}\label{sec:paired_ranks}

For mesh $j$, corruption $c$, and checkpoint $s$, let $r^T_{jcs}$ and $r^L_{jcs}$ be the archived ranks of the correct clean mesh under TUN and lifetime selection. Each pair uses the same corrupted observation. We verified equal selection counts in the stored query and gallery subsets and reproduced all per-checkpoint R@1, R@5 and mAP values for these two selectors. For each corruption we first average $1/r^T_{jcs}-1/r^L_{jcs}$ over the ten checkpoints for each mesh. The mean over the 400 meshes then gives the paired mAP difference in Table~\ref{tab:paired_ranks}. To describe variation across query identities, we resample 400 meshes with replacement 10,000 times (random seed 20261117). Each draw retains all conditions and checkpoints for a sampled mesh. The 2.5th and 97.5th percentiles give pointwise intervals conditional on the observed corruptions, frozen checkpoints and fixed clean gallery.

\begin{table}[htbp]
    \centering
    \caption{Paired CAD retrieval in the 400-mesh gallery. Higher, equal and lower refer to the correct source's position with TUN relative to lifetime selection at the same budget. Gained and lost count transitions into and out of first place. Counts are means per 400 queries over ten checkpoints. The final two rows also average over the indicated conditions. Brackets give 95\% mesh-bootstrap intervals for the mAP difference.}
    \label{tab:paired_ranks}
    \footnotesize
    \setlength{\tabcolsep}{3.5pt}
    \begin{tabular}{@{}lrrrcrr@{}}
        \toprule
        & \multicolumn{3}{c}{Correct-source rank} & & \multicolumn{2}{c}{First place} \\
        \cmidrule(lr){2-4}\cmidrule(l){6-7}
        Query condition & Higher & Equal & Lower & $\Delta$\,mAP [95\% interval] & Gained & Lost \\
        \midrule
        Density 25\%     & 14.3 & 305.5 & 80.2  & $-0.0020\;[-0.0072,\,0.0033]$ & 2.3 & 3.0 \\
        Density 50\%     & 9.9  & 344.9 & 45.2  & $-0.0032\;[-0.0098,\,0.0029]$ & 2.0 & 3.5 \\
        Jitter 0.5\%     & 31.2 & 231.2 & 137.6 & $+0.0023\;[-0.0023,\,0.0074]$ & 2.1 & 0.3 \\
        Jitter 1\%       & 27.2 & 177.1 & 195.7 & $-0.0003\;[-0.0011,\,0.0002]$ & 0.0 & 0.0 \\
        Non-uniform      & 42.5 & 262.3 & 95.2  & $-0.0081\;[-0.0221,\,0.0050]$ & 10.1 & 9.9 \\
        Outliers 1\%     & 179.0 & 155.0 & 66.0 & $+0.2331\;[0.1978,\,0.2696]$ & 93.5 & 3.0 \\
        Outliers 5\%     & 157.1 & 133.2 & 109.7 & $+0.0958\;[0.0721,\,0.1209]$ & 31.9 & 1.7 \\
        Partial 20\%     & 18.3 & 295.6 & 86.1 & $-0.0124\;[-0.0252,\,-0.0002]$ & 3.3 & 8.5 \\
        Partial 30\%     & 20.9 & 273.9 & 105.2 & $-0.0126\;[-0.0260,\,0.0006]$ & 5.4 & 6.8 \\
        \midrule
        All nine         & 55.6 & 242.1 & 102.3 & $+0.0325\;[0.0260,\,0.0392]$ & 16.7 & 4.1 \\
        Seven non-outlier & 23.5 & 270.1 & 106.5 & $-0.0052\;[-0.0097,\,-0.0011]$ & 3.6 & 4.6 \\
        \bottomrule
    \end{tabular}
\end{table}
\begin{figure}[htbp]
	\centering
	\includegraphics[width=\linewidth]{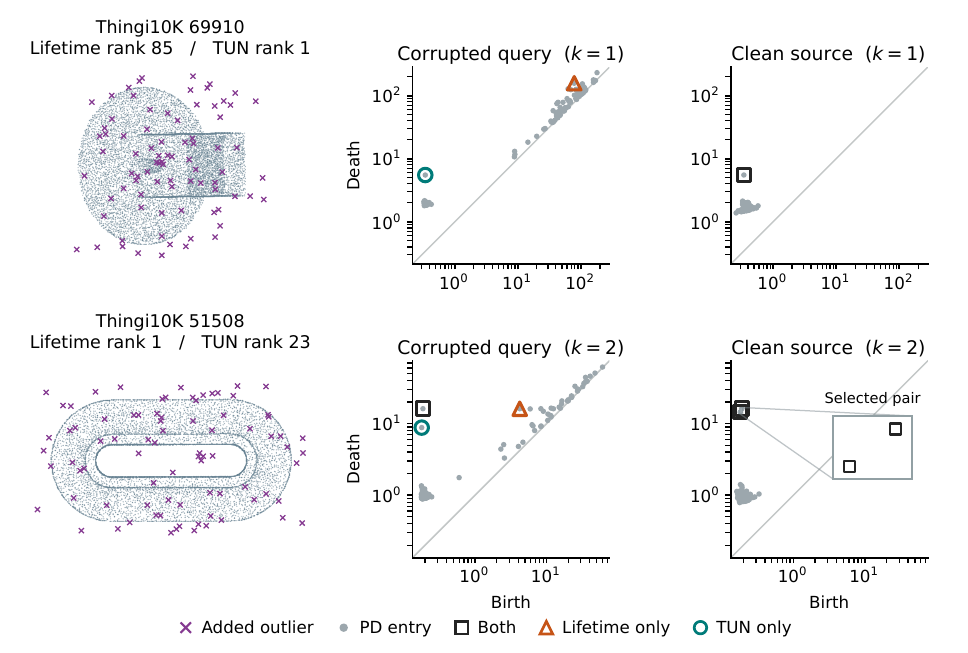}
	\caption{Two Thingi10K queries with 1\% outliers at checkpoint 0. Left, surface samples projected onto their first two principal axes, with injected points marked by crosses. Middle and right, the 100 retained query and clean-source $H_1$ intervals in native squared-radius units. The diagrams share logarithmic axis limits within each row. Squares mark intervals selected by both methods. The inset enlarges the two selected clean intervals for mesh 51508. Ranks use all 400 clean meshes.}
	\Description{For each mesh, the point cloud appears beside log-scale query and source diagrams. Purple crosses mark outliers. Squares indicate shared selections, triangles lifetime-only selections, and circles TUN-only selections. The correct-source ranks change from 85 to 1 and from 1 to 23.}
	\label{fig:retrieval_examples}
\end{figure}

Across the nine conditions, more ranks fall than rise (102.3 versus 55.6 per 400 queries). Reciprocal rank gives greater weight to changes near first place, where TUN gains 16.7 correct matches and loses 4.1 on average. At 1\% outliers, TUN improves 179 of 400 ranks per checkpoint on average, leaves 155 unchanged, and lowers 66. Across checkpoints, 111 meshes rank higher with TUN at all ten checkpoints, 35 rank lower at all ten, and 104 keep the same rank under both methods at every checkpoint. The archived ranking breaks equal distances by gallery order. Requiring the true source to be strictly closer than every other mesh gives R@1 differences of $+0.22625$ and $+0.07550$ at 1\% and 5\% outliers, respectively, the same differences as under the recorded tie rule.

\subsection{How interval choices change the retrieved source}

At matched counts, the identity of the selected intervals can change the retrieved source. Figure~\ref{fig:retrieval_examples} follows one first-place gain and one loss at checkpoint 0 under 1\% outliers. The two selectors retain the same counts and choose the same intervals on the correct clean source in both examples.
In the upper example, TUN selects the 45th longest query interval, whose birth and death values closely match the selected clean interval. Its correct-source rank improves from 85 to 1. In the lower example, replacing the second-longest query interval with the fifth-longest raises the distance to the correct source and moves its rank from 1 to 23. Appendix~\ref{app:retrieval_cases} provides the coordinates, distances, and the within-group median rule used to choose the cases.

\section{Conclusion}\label{sec:conclusion}

Choosing source-loop entries requires identifying intervals even when their order differs from lifetime ranking. Our benchmark records this choice through reviewed labels on four geometric categories. TUN combines diagram context with a conditional point-cloud correction to predict the full positive set, achieving higher mean F1 than its PD-only initialization through fewer false positives. At matched selection counts, its selected intervals also improve CAD source-identity retrieval under 1\% and 5\% added outliers.

A natural extension is to connect each candidate interval to a region of the point cloud or a representative cycle, especially for short intervals. Extending the task to $H_2$ would require source cavity counts and reviewed labels for individual intervals. Images and voxel grids provide further geometric inputs for this formulation.

\section*{Software}
The code of this paper is available at \url{https://github.com/ZJUCAGD/TopologyUnderstandingNet}.

\section*{Acknowledgement}
This work is supported in part by National Natural Science Foundation of China under Grant Nos. 12671457, 62272406.

\bibliographystyle{ACM-Reference-Format}

\bibliography{BIB}

\clearpage

\appendix

\counterwithin{figure}{section}

\counterwithin{table}{section}

\numberwithin{equation}{section}

\section{Source Geometries and Sampling}\label{app:benchmark}

Source shapes have first Betti number below 100, and the sampled point clouds range from hundreds to 100,000 points. We truncate or pad each PD to 100 entries and subsample or pad each cloud to 4,096 points. Masks exclude padding from attention, pooling, loss, and evaluation.

\begin{figure}[htbp]
    \centering
    \includegraphics[width=1.0\linewidth]{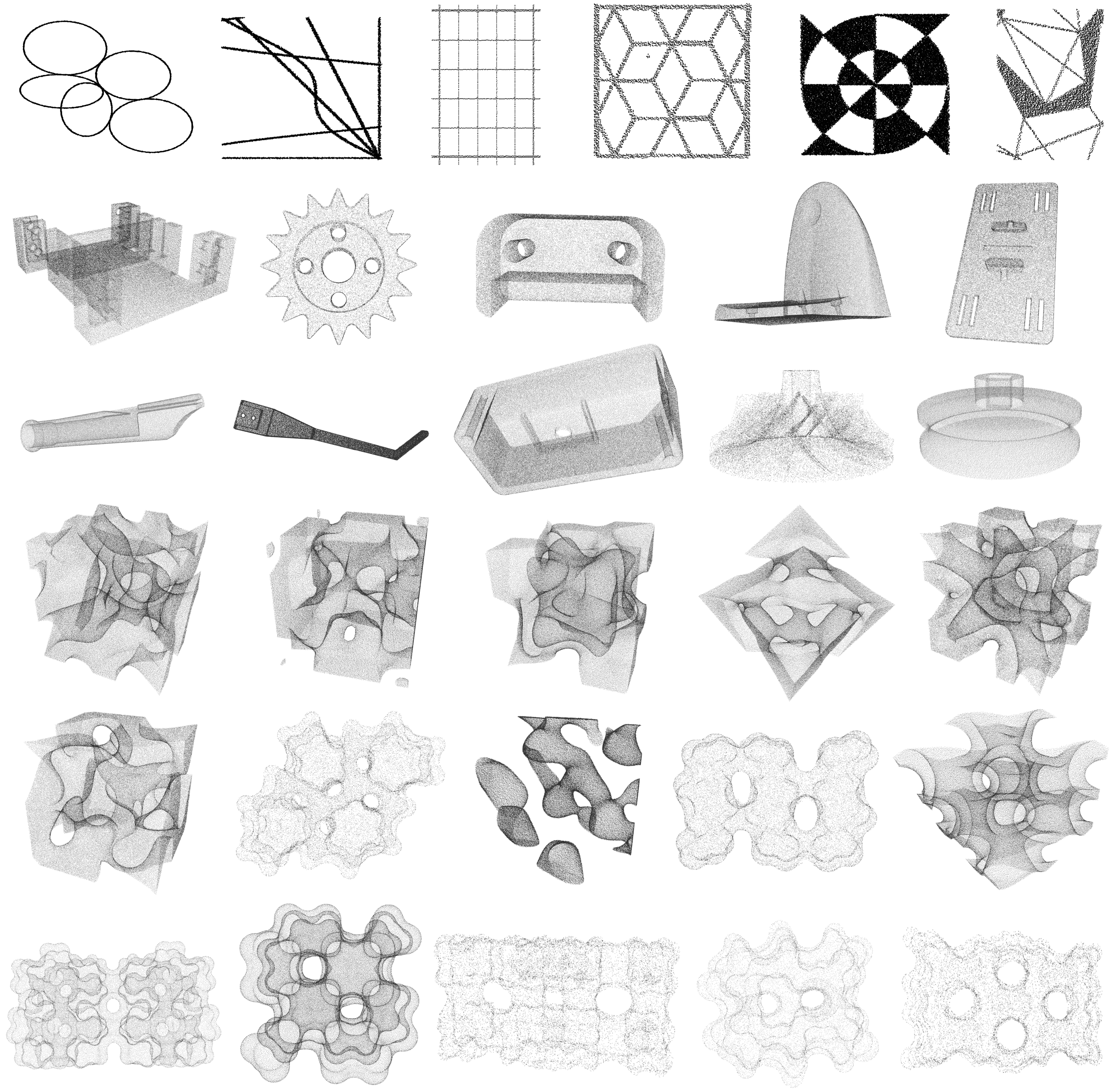}
    \caption{Examples of planar, CAD, TPMS, and zeolite point clouds used in the benchmark, grouped by category from top to bottom.}
    \Description{Grayscale renderings show examples in four groups. From top to bottom these are planar outlines, CAD objects, periodic minimal surfaces, and zeolite frameworks.}
    \label{fig:PC}
\end{figure}

For each instance, we use Poisson-disk sampling on the underlying geometric model and add a bounded point perturbation when specified. We then construct the alpha complex and compute its 1D persistence diagram with GUDHI. Figure~\ref{fig:PC} shows examples from all four categories.

\begin{figure}[htbp]
    \centering
    \includegraphics[width=0.9\linewidth]{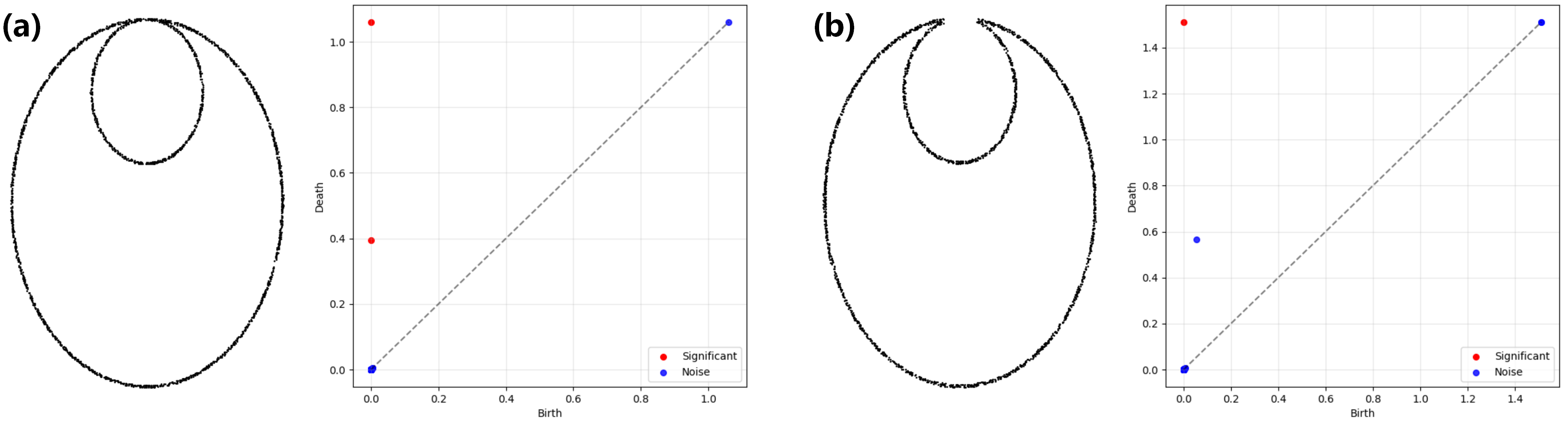}
    \caption{Two visually similar planar 1-PDs with different reviewed positive sets. The panels show the diagrams and labels for two distinct source shapes.}
    \Description{Each of two planar point clouds is paired with its birth--death plot. The left plot has two reviewed positives in red and the right plot has one. Negative entries appear in blue.}
    \label{fig:counter}
\end{figure}

Figure~\ref{fig:counter} pairs two visually similar planar 1-PDs with distinct reviewed positive sets. The corresponding source geometries determine the entries selected in the review.

\FloatBarrier

\section{Point-level Confusion Counts}\label{app:classification}
Selecting the longest intervals at the reviewed count gives F1 of $0.7990$ on TPMS and $0.9835$ on CAD, where the review changes fewer selections (Section~\ref{sec:label_audit}). The two count-free rules show different error profiles. The 2-means rule attains $0.3320$ unweighted mean F1 and produces many false positives. CS(0.5) yields precision above $0.97$ on Planar and CAD but recall below $0.15$. It predicts no positive entries on TPMS or Zeolite.

Among the point-wise learned references, the RBF SVM is stronger than the MLP on every category and reaches $0.9133$ unweighted mean F1. Absolute birth--death location carries substantial predictive signal, particularly on TPMS ($0.9653$), while TUN is higher by $0.0333$--$0.1234$ across categories. The MLP varies more across runs, with category-wise standard deviations of $0.0923$--$0.1502$.

Relative to its PD-only initialization, TUN gains F1 in all ten paired runs (four-category mean gain $0.0038$). False positives fall from $113.7$ to $25.0$ per run and false negatives rise from $20.0$ to $33.7$ (Tables~\ref{tab:confusion_counts} and~\ref{tab:confusion_counts_2}).

Tables~\ref{tab:confusion_counts} and~\ref{tab:confusion_counts_2} report the error composition underlying Table~\ref{tab:main_results}. Among the four set models, TUN yields the fewest false positives in every category. DeepSets and Set Transformer attain slightly higher recall on CAD, whereas TUN reduces false positives from $15.5$ for DeepSets to $8.1$; this precision--recall trade-off accounts for the $0.0010$ CAD F1 difference. On the other three categories, the same false-positive suppression accompanies the highest F1-score.

\begin{table}[htbp]
    \caption{Point-level confusion counts for Planar and CAD. Deterministic rules are evaluated once; values for learning methods are averaged over ten runs. Bold denotes the best value in each metric column: higher is better for TP/TN and lower is better for FP/FN.}
    \label{tab:confusion_counts}
    \centering
    \footnotesize
    \setlength{\tabcolsep}{3pt}
    \begin{tabular}{lrrrrrrrr}
        \toprule
        & \multicolumn{4}{c}{\textbf{Planar}} & \multicolumn{4}{c}{\textbf{CAD}} \\
        \cmidrule(lr){2-5} \cmidrule(lr){6-9}
        \textbf{Method} & \textbf{TP} & \textbf{TN} & \textbf{FP} & \textbf{FN} & \textbf{TP} & \textbf{TN} & \textbf{FP} & \textbf{FN} \\
        \midrule
        2-means & 1182 & 18483 & 5131 & 204 & 1388 & 34323 & 13856 & 433 \\
        CS ($0.5$) & 201 & 23609 & 5 & 1185 & 123 & \textbf{48176} & \textbf{3} & 1698 \\
        \midrule
        SVM & 1254.4 & 23555.5 & 58.5 & 131.6 & 1634.1 & 47878.0 & 301.0 & 186.9 \\
        MLP & 492.5 & \textbf{23609.2} & \textbf{4.8} & 893.5 & 306.8 & 48066.6 & 112.4 & 1514.2 \\
        \midrule
        DeepSets & \textbf{1386.0} & 23578.5 & 35.5 & \textbf{0.0} & 1816.3 & 48163.5 & 15.5 & 4.7 \\
        Set Transformer & 1385.7 & 23571.8 & 42.2 & 0.3 & \textbf{1820.4} & 48149.4 & 29.6 & \textbf{0.6} \\
        PD-Encoder only & 1378.6 & 23587.4 & 26.6 & 7.4 & 1815.9 & 48140.6 & 38.4 & 5.1 \\
        \textbf{TUN} & 1381.8 & 23601.8 & 12.2 & 4.2 & 1805.5 & 48170.9 & 8.1 & 15.5 \\
        \bottomrule
    \end{tabular}
\end{table}

\begin{table}[htbp]
    \caption{Point-level confusion counts for TPMS and Zeolite. Deterministic rules are evaluated once; values for learning methods are averaged over ten runs. TPMS corresponds to the porous-structure split in the result files. Bold denotes the best value in each metric column: higher is better for TP/TN and lower is better for FP/FN.}
    \label{tab:confusion_counts_2}
    \centering
    \footnotesize
    \setlength{\tabcolsep}{3pt}
    \begin{tabular}{lrrrrrrrr}
        \toprule
        & \multicolumn{4}{c}{\textbf{TPMS}} & \multicolumn{4}{c}{\textbf{Zeolite}} \\
        \cmidrule(lr){2-5} \cmidrule(lr){6-9}
        \textbf{Method} & \textbf{TP} & \textbf{TN} & \textbf{FP} & \textbf{FN} & \textbf{TP} & \textbf{TN} & \textbf{FP} & \textbf{FN} \\
        \midrule
        2-means & 3449 & 32052 & 12561 & 1938 & 4564 & 11524 & 7766 & 146 \\
        CS ($0.5$) & 0 & \textbf{44613} & \textbf{0} & 5387 & 0 & \textbf{19290} & \textbf{0} & 4710 \\
        \midrule
        SVM & 5195.6 & 44431.2 & 181.8 & 191.4 & 4702.2 & 18111.6 & 1178.4 & 7.8 \\
        MLP & 1107.0 & 44535.5 & 77.5 & 4280.0 & 194.5 & 19289.7 & 0.3 & 4515.5 \\
        \midrule
        DeepSets & 5378.8 & 44541.2 & 71.8 & 8.2 & 4707.4 & 19280.7 & 9.3 & 2.6 \\
        Set Transformer & \textbf{5383.6} & 44563.3 & 49.7 & \textbf{3.4} & \textbf{4709.9} & 19277.4 & 12.6 & \textbf{0.1} \\
        PD-Encoder only & 5379.8 & 44581.9 & 31.1 & 7.2 & 4709.7 & 19272.4 & 17.6 & 0.3 \\
        \textbf{TUN} & 5375.6 & 44609.6 & 3.4 & 11.4 & 4707.4 & 19288.7 & 1.3 & 2.6 \\
        \bottomrule
    \end{tabular}
\end{table}

\FloatBarrier

\section{Detailed CAD Retrieval Results}\label{app:retrieval}

\subsection{Interval selections and example construction}\label{app:retrieval_cases}

The two selectors choose different intervals for 281.6 of the 400 queries on average at 1\% outliers. These pairs contribute $+0.23329$ to the overall mAP difference and the pairs with identical query selections contribute $-0.00016$, each sum divided by all 4,000 query--checkpoint pairs. The corresponding contributions at 5\% outliers are $+0.09566$ and $+0.00014$. Gallery subsets also differ for 48.6 of the 400 clean meshes on average, so the first contribution includes changes on both sides of the retrieval comparison.

Figure~\ref{fig:retrieval_examples} shows one first-place gain and one loss for checkpoint 0. Within each transition group we order queries by absolute rank change, breaking ties by source filename, and take the upper median. The groups contain 106 gains and one loss. We replayed the 400 assignment distances under each selector from the stored intervals for both examples.

For mesh 69910, both selectors keep one query interval. The longest interval is $(78.27,157.42)$ in squared-radius coordinates. TUN selects the 45th longest, $(0.341,5.523)$, close to the clean interval $(0.344,5.528)$ selected by both methods. The distance to the correct mesh falls from $39.660$ to $0.0062$ and its rank improves from 85 to 1. For mesh 51508, both methods retain the longest query interval. TUN replaces the second longest with the fifth longest. Both selectors retain the same two clean-source intervals. The substitution raises the distance to the source from $4.370$ to $5.639$ and its rank falls from 1 to 23. Across the ten checkpoints, mesh 69910 gains first place six times and mesh 51508 loses it all ten times.

\subsection{Results by corruption}\label{app:retrieval_results}

Table~\ref{tab:cad_retrieval_detailed} gives the results for each of the nine corruptions in the full gallery. The aggregate results in Table~\ref{tab:cad_retrieval_protocols} average equally over these conditions.

\begin{table}[htbp]
    \caption{Full-gallery CAD retrieval for each query corruption. Values are means and sample standard deviations over ten seeds. Random-matched is first averaged over five subset draws. Bold marks the largest displayed mean for each condition and metric.}
    \label{tab:cad_retrieval_detailed}
    \centering
    \footnotesize
    \setlength{\tabcolsep}{4pt}
    \begin{tabular}{llccc}
        \toprule
        \textbf{Corruption} & \textbf{Selection} & \textbf{R@1} & \textbf{R@5} & \textbf{mAP} \\
        \midrule
        Density (25\%) & Count-only & $0.0115\!\pm\!0.0036$ & $0.0530\!\pm\!0.0075$ & $0.0449\!\pm\!0.0052$ \\
        & Random-matched & $0.0181\!\pm\!0.0023$ & $0.0402\!\pm\!0.0042$ & $0.0368\!\pm\!0.0028$ \\
        & Top-persistence@$k_{\mathrm{TUN}}$ & $\mathbf{0.4008\!\pm\!0.0503}$ & $\mathbf{0.5677\!\pm\!0.0669}$ & $\mathbf{0.4773\!\pm\!0.0574}$ \\
        & TUN & $0.3990\!\pm\!0.0473$ & $0.5655\!\pm\!0.0629$ & $0.4753\!\pm\!0.0537$ \\
        \addlinespace
        Density (50\%) & Count-only & $0.0180\!\pm\!0.0045$ & $0.0853\!\pm\!0.0154$ & $0.0666\!\pm\!0.0075$ \\
        & Random-matched & $0.0273\!\pm\!0.0042$ & $0.0616\!\pm\!0.0108$ & $0.0558\!\pm\!0.0060$ \\
        & Top-persistence@$k_{\mathrm{TUN}}$ & $\mathbf{0.5723\!\pm\!0.0413}$ & $\mathbf{0.7080\!\pm\!0.0464}$ & $\mathbf{0.6365\!\pm\!0.0429}$ \\
        & TUN & $0.5685\!\pm\!0.0417$ & $0.7032\!\pm\!0.0464$ & $0.6333\!\pm\!0.0430$ \\
        \addlinespace
        Jitter (0.5\% bbox) & Count-only & $0.0158\!\pm\!0.0035$ & $0.0698\!\pm\!0.0070$ & $0.0565\!\pm\!0.0042$ \\
        & Random-matched & $0.0252\!\pm\!0.0039$ & $0.1047\!\pm\!0.0059$ & $0.0782\!\pm\!0.0052$ \\
        & Top-persistence@$k_{\mathrm{TUN}}$ & $0.0638\!\pm\!0.0056$ & $\mathbf{0.2533\!\pm\!0.0089}$ & $0.1609\!\pm\!0.0052$ \\
        & TUN & $\mathbf{0.0683\!\pm\!0.0059}$ & $0.2513\!\pm\!0.0079$ & $\mathbf{0.1632\!\pm\!0.0059}$ \\
        \addlinespace
        Jitter (1\% bbox) & Count-only & $\mathbf{0.0087\!\pm\!0.0029}$ & $\mathbf{0.0457\!\pm\!0.0079}$ & $\mathbf{0.0390\!\pm\!0.0043}$ \\
        & Random-matched & $0.0054\!\pm\!0.0010$ & $0.0328\!\pm\!0.0021$ & $0.0308\!\pm\!0.0008$ \\
        & Top-persistence@$k_{\mathrm{TUN}}$ & $0.0025\!\pm\!0.0000$ & $0.0273\!\pm\!0.0014$ & $0.0361\!\pm\!0.0005$ \\
        & TUN & $0.0025\!\pm\!0.0000$ & $0.0273\!\pm\!0.0014$ & $0.0358\!\pm\!0.0005$ \\
        \addlinespace
        Non-uniform ($w\propto A^2$) & Count-only & $0.0145\!\pm\!0.0044$ & $0.0678\!\pm\!0.0063$ & $0.0551\!\pm\!0.0033$ \\
        & Random-matched & $0.0452\!\pm\!0.0050$ & $0.1076\!\pm\!0.0073$ & $0.0857\!\pm\!0.0063$ \\
        & Top-persistence@$k_{\mathrm{TUN}}$ & $0.4348\!\pm\!0.0187$ & $\mathbf{0.5822\!\pm\!0.0279}$ & $\mathbf{0.5074\!\pm\!0.0223}$ \\
        & TUN & $\mathbf{0.4353\!\pm\!0.0225}$ & $0.5657\!\pm\!0.0263$ & $0.4992\!\pm\!0.0247$ \\
        \addlinespace
        Outliers (1\%) & Count-only & $0.0457\!\pm\!0.0086$ & $0.1710\!\pm\!0.0188$ & $0.1232\!\pm\!0.0128$ \\
        & Random-matched & $0.0873\!\pm\!0.0066$ & $0.2274\!\pm\!0.0161$ & $0.1629\!\pm\!0.0111$ \\
        & Top-persistence@$k_{\mathrm{TUN}}$ & $0.0795\!\pm\!0.0074$ & $0.1585\!\pm\!0.0081$ & $0.1251\!\pm\!0.0076$ \\
        & TUN & $\mathbf{0.3058\!\pm\!0.0392}$ & $\mathbf{0.4120\!\pm\!0.0408}$ & $\mathbf{0.3583\!\pm\!0.0383}$ \\
        \addlinespace
        Outliers (5\%) & Count-only & $0.0265\!\pm\!0.0046$ & $0.1095\!\pm\!0.0118$ & $0.0847\!\pm\!0.0053$ \\
        & Random-matched & $0.0073\!\pm\!0.0018$ & $0.0308\!\pm\!0.0024$ & $0.0285\!\pm\!0.0021$ \\
        & Top-persistence@$k_{\mathrm{TUN}}$ & $0.0107\!\pm\!0.0012$ & $0.0325\!\pm\!0.0035$ & $0.0311\!\pm\!0.0013$ \\
        & TUN & $\mathbf{0.0863\!\pm\!0.0128}$ & $\mathbf{0.1585\!\pm\!0.0203}$ & $\mathbf{0.1269\!\pm\!0.0155}$ \\
        \addlinespace
        Partial (20\% removed) & Count-only & $0.0105\!\pm\!0.0033$ & $0.0452\!\pm\!0.0070$ & $0.0499\!\pm\!0.0032$ \\
        & Random-matched & $0.0952\!\pm\!0.0059$ & $0.2542\!\pm\!0.0133$ & $0.1779\!\pm\!0.0085$ \\
        & Top-persistence@$k_{\mathrm{TUN}}$ & $\mathbf{0.4490\!\pm\!0.0245}$ & $\mathbf{0.5760\!\pm\!0.0272}$ & $\mathbf{0.5121\!\pm\!0.0253}$ \\
        & TUN & $0.4360\!\pm\!0.0207$ & $0.5633\!\pm\!0.0254$ & $0.4996\!\pm\!0.0226$ \\
        \addlinespace
        Partial (30\% removed) & Count-only & $0.0107\!\pm\!0.0026$ & $0.0393\!\pm\!0.0055$ & $0.0417\!\pm\!0.0038$ \\
        & Random-matched & $0.0707\!\pm\!0.0045$ & $0.2008\!\pm\!0.0105$ & $0.1403\!\pm\!0.0064$ \\
        & Top-persistence@$k_{\mathrm{TUN}}$ & $\mathbf{0.3430\!\pm\!0.0226}$ & $\mathbf{0.4695\!\pm\!0.0268}$ & $\mathbf{0.4067\!\pm\!0.0243}$ \\
        & TUN & $0.3395\!\pm\!0.0214$ & $0.4383\!\pm\!0.0203$ & $0.3940\!\pm\!0.0217$ \\
        \bottomrule
    \end{tabular}
\end{table}

\end{document}